\documentclass[runningheads]{llncs}
\usepackage{graphicx}

\usepackage{tikz}
\usepackage{comment}
\usepackage{amsmath,amssymb} 
\usepackage{color}

\usepackage[accsupp]{axessibility}  

\usepackage{stmaryrd}
\usepackage{physics}
\DeclareMathOperator*{\argmax}{arg\,max}

\begin{document}
\pagestyle{headings}
\mainmatter
\def\ECCVSubNumber{14}  

\title{Consistency Regularization for \\Domain Adaptation} 

\titlerunning{Consistency Regularization for Domain Adaptation}
%
\author{Kian Boon Koh\inst{1,2}\and 
Basura~Fernando\inst{1,2}}
%
%
\institute{Institute of High Performance Computing, A*STAR, Singapore \and Centre for Frontier AI Research, A*STAR, Singapore\\
\email{kianboonkoh@gmail.com}\\
\email{fernando\_basura@ihpc.a-star.edu.sg}}
\maketitle

\begin{abstract}
Collection of real world annotations for training semantic segmentation models is an expensive process. Unsupervised domain adaptation (UDA) tries to solve this problem by studying how more accessible data such as synthetic data can be used to train and adapt models to real world images without requiring their annotations. Recent UDA methods applies self-learning by training on pixel-wise classification loss using a student and teacher network. In this paper, we propose the addition of a consistency regularization term to semi-supervised UDA by modelling the inter-pixel relationship between elements in networks' output. We demonstrate the effectiveness of the proposed consistency regularization term by applying it to the state-of-the-art DAFormer framework and improving mIoU19 performance on the GTA5 to Cityscapes benchmark by 0.8 and mIou16 performance on the SYNTHIA to Cityscapes benchmark by 1.2.
\end{abstract}

\section{Introduction}
\label{sec:intro}
Semantic segmentation is a task which requires a lot of pixel level annotations and obtaining these annotations is expensive and time consuming. 
To overcome this issue, one solution is to obtain annotations from synthetic data such as Games (e.g. GTA5) and train models on these synthetic data for semantic segmentation.
However, the problem is that even if modern synthetic data is near photo realistic, still there is a distribution mismatch between the synthetic data and real images.
One solution is to develop models that can overcome this distribution mismatch between models that are trained on synthetic data and real data which is the topic of unsupervised domain adaptation (UDA)~\cite{Fernando2017,fernando2013unsupervised,herath2019min,ganin2015unsupervised,pan2010domain,ben2006analysis,tzeng2017adversarial,DBLP:journals/corr/FernandoHST14}.

UDA for semantic segmentation has made significant progress in recent years.
One of the most recent method called DAFormer~\cite{DBLP:journals/corr/abs-2111-14887} obtained massive improvement over prior methods by using a Transformer architecture and self-training. However one of the challenges in self-training is that generated pseudo labels can be wrong and that may result in poor transfer of information from source domain to the target domain. Therefore, it is needed to further regularize the self-training learning process.

In this work, we present a new consistency regularization method based on correlation between pixel-wise class predictions.
We enforce two models (teacher and student) to have similar inter-pixel similarity structure and by doing so we regularize the self-training process.
This helps to improve the generalization of the student network as well as the teacher network allowing better transfer of information from the source domain to the target domain. We demonstrate its effectiveness by applying it to DAFormer and improving mIoU19 performance on the GTA5 to Cityscapes benchmark by 0.8 and mIou16 performance on the SYNTHIA to Cityscapes benchmark by 1.2.
Implementation of our proposed method is available at our GitHub repository\footnote{\url{https://github.com/kw01sg/CRDA}}.

\section{Related Work}

\subsection{Unsupervised Domain Adaptation}

Domain adaptation is a field of techniques that aims to solve the domain shift problem, when data distributions experience change between datasets. UDA is a subset of the domain adaptation field that aims to utilize a labeled source domain to learn a model that performs well on an unlabeled target domain. 
Recent UDA methods can be grouped into either adversarial training or self-supervised learning (SSL) approaches. Adversarial training methods aim to reduce source and target distribution mismatch by aligning distributions at either the pixel \cite{Hoffman_cycada2017,8953759,DBLP:journals/corr/BousmalisSDEK16} or intermediate feature level \cite{DBLP:journals/corr/TzengHSD17,hoffman2016fcns} using a generative adversarial network (GAN). 

SSL methods allow models to be trained directly on the target domain by generating pseudo labels from the target domain. Recent advances focuses on improving the quality of pseudo labels using various approaches, such as using representative prototypes \cite{zhang2021prototypical} or using more complex, Transformer-based network architecture \cite{DBLP:journals/corr/abs-2111-14887}. 
It is also possible for methods to adopt a hybrid approach and use both adversarial training and SSL. Li et al. does so in their bidirectional learning framework \cite{8954260}. Adversarial training is first used to obtain an image-to-image translation model and a segmentation model. Target domain pseudo labels are then generated from high confidence predictions, which are then subsequently used to fine tune the segmentation model. The improved segmentation model can then be used in the first adversarial stage to form a close loop.

\subsection{Semantic Segmentation}

Early methods on semantic segmentation problems were largely based on Fully Convolutional Network (FCN) \cite{7478072}, which typically follows an encoder-decoder architecture \cite{Badrinarayanan2017SegNetAD,10.1007/978-3-319-24574-4_28}. Further improvements were made by using dilated convolutions to overcome the loss of spatial resolution \cite{YuKoltun2016}, and pyramid pooling \cite{8100143,7913730} to enhance capturing of contextual information. 
Recent success of attention-based Transformers \cite{NIPS2017_3f5ee243} in natural language processing has seen adaptations of Transformers for image segmentation \cite{liu2021Swin,xie2021segformer} that were able to obtain state-of-the-art results.  

\subsection{Consistency Regularization}

Consistency regularization is a regularization technique used to encourage networks to make consistent predictions that are invariant to perturbations. Tarvainen and Valpola improved model performance on the image classification problem by using a student and teacher network pair in their Mean Teacher model \cite{DBLP:journals/corr/TarvainenV17}, where the weights of the teacher network are an exponential moving average (EMA) of the student network. Consistent predictions between the two networks are then promoted by optimizing a consistency loss between their predictions. 
Interpolation consistency training by Verma et al. \cite{ijcai2019-504} combines mixup \cite{zhang2018mixup} and the Mean Teacher model \cite{DBLP:journals/corr/TarvainenV17} to implement consistency regularization. During training, unlabelled samples are interpolated to create an augmented sample. Predictions by the student network on the augmented sample are then optimized to be consistent with interpolated predictions by the teacher network on the original non-interpolated samples.
Kim et al. \cite{DBLP:journals/corr/abs-2001-04647} uses cosine similarity in their consistency regularization method for semantic segmentation. They propose a structured consistency loss that optimizes predictions to be consistent in not only pixel-wise classification, but also inter-pixel relationship. 

\section{Our Method}

Given source domain images $x_S \in X_S$ with their annotations (labels) $y_S \in Y_S$ and target domain images $x_T \in X_T$ without annotations (labels), we want to learn a network $h$ that can correctly predict the annotations for target images $X_T$ denoted by $\hat{Y}_T$. Typically, there is a mismatch in the joint probability distributions of source domain data $P(X_S,Y_S)$ and the target domain data $P(X_T,Y_T)$.
Due to this mismatch or the gap between source and target domains, an image segmentation model $h$ that is trained on the source data usually results in a low performance on target images. 
One common solution to address this issue is to use self-training as also done in the prior works such as DAFormer \cite{DBLP:journals/corr/abs-2111-14887}.
However, semi-supervised self-training methods could easily over-fit to source distribution and could generate inconsistent or wrong pseudo labels for the target domain images.  
To overcome this limitation, we propose the addition of consistency regularization to the DAFormer \cite{DBLP:journals/corr/abs-2111-14887} framework during model training to further improve model performance.
Next, we explain the overall training framework.

\subsection{Overall  Training}

Overall training of the network is composed of three components: supervised training using source images, self-training using target images, and consistency regularization. Total loss $\mathcal{L}_{total}$ is given as
\begin{equation}
    \mathcal{L}_{total} = \mathcal{L}_S + \mathcal{L}_T + \lambda_c \mathcal{L}_C
\end{equation} 
where $\mathcal{L}_S$ is supervised cross entropy loss using source images, $\mathcal{L}_T$ is self-trained cross entropy loss using pseudo labels, $\mathcal{L}_C$ is our consistency regularization term, and $\lambda_c$ is a parameter we use to weigh $\mathcal{L}_C$. The following sections will present each of the losses in detail.

\subsubsection{Supervised Training}

Supervised training on the source domain is conducted using cross entropy loss for semantic segmentation. For a source image $x_S$ and its annotation $y_S$, $\mathcal{L}_S$ can be defined as

\begin{equation}
    \mathcal{L}_S(x_S, y_S) = - \frac{1}{H W} \sum_{j=1}^{H \times W} \sum_{c=1}^C y_S^{(j,c)} \log h(x_S)^{(j,c)}
\end{equation}
where $C$ is the number of classes and $H$ and $W$ are the height and width of the segmentation output. The notation $y_S^{(j,c)}$ denotes the presence of class $c$ at pixel location $j$ (1 if present and 0 if not). Similarly, $h(x_S)^{(j,c)}$ denotes the predicted score for class $c$ at pixel location $j$ using model $h$ for image $x_S$.

\subsubsection{Self-Training}

Self-training uses a teacher network $f(;\phi)$ to produce pseudo labels on which the student network $h(;\theta)$ will be trained on. For a target image $x_T$, its pseudo label $p_T$ is formally defined as

\begin{equation}
    p^{(j,c)}_T = \llbracket c = \argmax_{c'} f(x_T;\phi)^{(j,c')} \rrbracket
\label{eq.label}
\end{equation}
where $\llbracket \cdot \rrbracket$ denotes the Iverson bracket.

We follow the Mean Teacher model \cite{DBLP:journals/corr/TarvainenV17} where the weights of the teacher network $f(;\phi)$ are the EMA of the weights of the student network $h(;\theta)$ after each training step $t$. The EMA weights used by the teacher model at training step $t$ is formally defined as
\begin{equation}
    \phi_{t+1} = \alpha \phi_{t} + (1 - \alpha) \theta_t
\end{equation}
where $\phi_{t+1}$ is the EMA of successive weights and $\alpha$ is a smoothing coefficient hyperparameter. It should also be noted that no gradients will be backpropagated into the teacher network from the student network.

A confidence estimate for the pseudo labels, defined as the ratio of pixels with maximum softmax probability exceeding a pre-defined threshold $\tau$, is also used in the self-training loss. For a target image $x_T$, its confidence estimate $q_T$ is formally defined as

\begin{equation}
    q_T = \frac{\sum^{H \times W}_{j=1}[\max_{c'}f(x_T; \phi)^{j,c'} > \tau]}{HW}
\end{equation}

Self-training loss of the student network $\mathcal{L}_T$ for a target image $x_T$ can thus be defined as

\begin{equation} \label{eq:self-training}
    \mathcal{L}_T(x_T) = - \frac{1}{H W} \sum_{j=1}^{H \times W} \sum_{c=1}^C q_T \times p^{(j,c)}_T \times \log h(x_T; \theta)^{(j,c)}
\end{equation}

We follow DAFormer's \cite{DBLP:journals/corr/abs-2111-14887} method of using non-augmented target images for the teacher network $f$ to generate pseudo labels and augmented targeted images to train the student network $h$ using Equation~\ref{eq:self-training}. We also follow their usage of color jitter, Gaussian blur, and ClassMix \cite{9423297} as data augmentations in our training process.

\subsubsection{Consistency Regularization} \label{consistency-regularization}

As mentioned in Mean Teacher \cite{DBLP:journals/corr/TarvainenV17}, cross entropy loss in Equation \ref{eq:self-training} between predictions of the student model and pseudo labels (which are predictions from the teacher model) can be considered as a form of consistency regularization. 
However, different from classification problems, semantic segmentation problems have a property where pixel-wise class predictions are correlated with each other. Thus, we propose to further enhance consistency regularization by focusing on this inter-pixel relationship. 
Inspired by the method of Kim et al. \cite{DBLP:journals/corr/abs-2001-04647}, we use the inter-pixel cosine similarity of networks' predictions on target images to regularize the model. Formally, we define the similarity between pixel $i$ and $j$ class predictions on a target image $x_T$ as
\begin{equation} \label{eq:similarity}
    a_{i,j} = \frac{\textbf{p}^T_i \textbf{p}_j}{\norm{\textbf{p}_i} \cdot \norm{\textbf{p}_j}}
\end{equation}
where $a_{i,j}$ represents the cosine similarity between the prediction vector of the \textit{i}th pixel and the prediction vector of the \textit{j}th pixel.
Note that the similarity between the probability vector $\textbf{p}_i$  and $\textbf{p}_j$ can also be computed using Kullback-Leibler (KL) divergence and cross entropy.
We investigate these options in Section \ref{inter-pixel-similarity}.
The consistency regularization term, $\mathcal{L}_C$ can then be defined as the mean squared error (MSE) between the student network's similarity matrix and the teacher network's similarity matrix
\begin{equation} \label{eq:consistency-regularization}
    \mathcal{L}_C = \frac{1}{(HW)^2} \sum^{H \times W}_{i=1} \sum^{H \times W}_{j=1} \norm{a^s_{i,j} - a^t_{i,j}}^2
\end{equation}
where $a^s_{i,j}$ is the similarity  obtained from the student network and $a^t_{i,j}$ is the similarity  obtained from the teacher network.
We also follow the method of Kim et al. \cite{DBLP:journals/corr/abs-2001-04647} to restrict the number of pixels used in the calculation of similarity matrices by performing a random sample of $N_{pair}$ pixels for comparison. Thus, the consistency regularization in Equation \ref{eq:consistency-regularization} is updated to the following equation

\begin{equation} \label{eq:consistency-regularization-with-n-pairs}
    \mathcal{L}_C = \frac{1}{(N_{pair})^2} \sum^{N_{pair}}_{i=1} \sum^{N_{pair}}_{j=1} \norm{a^s_{i,j} - a^t_{i,j}}^2
\end{equation}

This term $\mathcal{L}_C$ is particularly useful for domain adaptation as it helps to minimize the divergence between the source representation and the target representation by enforcing a structural consistency in the image segmentation task.

\section{Experiments}

\subsection{Implementation Details}

\subsubsection{Datasets}

We use the Cityscapes street scenes dataset \cite{Cordts2016Cityscapes} as our target domain. Cityscapes contains 2975 training and 500 validation images with resolution of 2048$\times$1024, and is labelled with 19 classes. 
For our source domain, we use the synthetic datasets GTA5 \cite{Richter_2016_ECCV} and SYNTHIA \cite{7780721}. GTA5 contains 24,966 images with resolution of 1914$\times$1052, and is labelled with the same 19 classes as Cityscapes. For compatibility, we use a variant of SYNTHIA that is labelled with 16 of the 19 Cityscapes classes. It contains 9,400 images with resolution of 1280$\times$760.
Following DAFormer \cite{DBLP:journals/corr/abs-2111-14887}, we resize images from Cityscapes to 1024$\times$512 pixels and images from GTA5 to 1280$\times$720 pixels before training.

\subsubsection{Network Architecture}

Our implementation is based on DAFormer \cite{DBLP:journals/corr/abs-2111-14887}.
Previous UDA methods mostly used DeepLabV2 \cite{deeplab} or FCN8s \cite{7478072} network architecture with ResNet \cite{7780459} or VGG \cite{simonyan2014} backbone as their segmentation model. DAFormer proposes an updated UDA network architecture based on Transformers that was able to achieve state-of-the-art performance. They hypothesized that self-attention is more effective than convolutions in fostering the learning of domain-invariant features.

\subsubsection{Training}

We follow DAFormer \cite{DBLP:journals/corr/abs-2111-14887} and train the network with AdamW \cite{loshchilov2018decoupled}, a learning rate of $\eta_{base} = 6 \times 10^{-5}$ for the encoder and $6 \times 10^{-4}$ for the decoder, a weight decay of 0.01, linear learning rate warmup with $t_{warm} = 1500$, and linear decay. Images are randomly cropped to $512 \times 512$ and trained for 40,000 iterations on a batch size of 2 on a NVIDIA GeForce RTX 3090.
We also adopt DAFormer's training strategy of rare class sampling and thing-class ImageNet feature distance to further improve results.
For hyperparameters used in self-training, we follow DAFormer and set $\alpha=0.99$ and $\tau=0.968$.
For hyperparameters used in consistency regularization, we set $N_{pair} = 512$, $\lambda_c = 1.0$ when calculating similarity using cosine similarity and $\lambda_c = 0.8 \times 10^{-3}$ when calculating similarity using KL divergence.
\subsection{Results}

\begin{table}
\centering
\caption{Comparison with other UDA methods on GTA5 to Cityscapes. Results for DAFormer and our method using cosine similarity are averaged over 6 random runs, while results for our method using KL Divergence are averaged over 3 random runs}
\resizebox{\textwidth}{!}{%
\begin{tabular}{l|ccccccccccccccccccc|cl}
Method   & Road          & Sidewalk      & Build.        & Wall          & Fence         & Pole          & Tr.Light      & Sign          & Veget.        & Terrain       & Sky           & Person        & Rider         & Car           & Truck         & Bus           & Train         & M.bike        & Bike          & mIoU19        &  \\
         \hline
BDL      & 91.0          & 44.7          & 84.2          & 34.6          & 27.6          & 30.2          & 36.0          & 36.0          & 85.0          & 43.6          & 83.0          & 58.6          & 31.6          & 83.3          & 35.3          & 49.7          & 3.3           & 28.8          & 35.6          & 48.5          &  \\
ProDA    & 87.8          & 56.0          & 79.7          & 46.3          & 44.8          & 45.6          & 53.5          & 53.5          & 88.6          & 45.2          & 82.1          & 70.7          & 39.2          & 88.8          & 45.5          & 59.4          & 1.0           & 48.9          & 56.4          & 57.5          &  \\
DAFormer      & 95.5          & 68.9          & 89.3          & \textbf{53.2} & 49.3          & 47.8          & \textbf{55.5} & 61.2          & 89.5          & 47.7          & 91.6          & 71.1          & 43.3          & 91.3          & 67.5          & 77.6          & 65.5          & 53.6          & 61.2          & 67.4          \\
Ours (Cosine) & 95.5          & 69.2          & \textbf{89.5} & 52.1          & \textbf{49.6} & 48.9          & 55.2          & \textbf{62.1} & 89.8          & 49.0          & 91.1          & \textbf{71.7} & \textbf{45.1} & \textbf{91.7} & \textbf{70.0} & 77.6          & 65.2          & \textbf{56.6} & 62.8          & 68.0          \\
Ours (KL)     & \textbf{96.1} & \textbf{71.6} & \textbf{89.5} & \textbf{53.2} & 48.6          & \textbf{49.5} & 54.7          & 61.1          & \textbf{90.0} & \textbf{49.4} & \textbf{91.7} & 70.7          & 44.0          & 91.6          & \textbf{70.0} & \textbf{78.1} & \textbf{68.9} & 55.1          & \textbf{62.9} & \textbf{68.2}
\end{tabular}%
}
\label{tab:gta_results}
\end{table}

\begin{table}
\centering
\caption{Comparison with other UDA methods on SYNTHIA to Cityscapes. Results for DAFormer and our method using cosine similarity are averaged over 6 random runs, while results for our method using KL Divergence are averaged over 3 random runs}
\resizebox{\textwidth}{!}{%
\begin{tabular}{l|cccccccccccccccc|cc}
Method   & Road          & Sidewalk      & Build.        & Wall          & Fence        & Pole          & Tr.Light      & Sign          & Veget.        & Sky           & Person        & Rider & Car           & Bus           & M.bike        & Bike          & mIoU16        & mIoU13        \\
         \hline
BDL      & 86.0          & 46.7          & 80.3          & -             & -            & -             & 14.1          & 11.6          & 79.2          & 81.3          & 54.1          & 27.9  & 73.7          & 42.2          & 25.7          & 45.3          & -             & 51.4          \\
ProDA    & 87.8          & 45.7          & 84.6          & 37.1          & 0.6          & 44.0          & 54.6          & 37.0          & 88.1          & 84.4          & 74.2          & 24.3  & 88.2          & 51.1          & 40.5          & 40.5          & 55.5          & 62.0          \\
DAFormer      & 80.5          & 37.6          & 87.9          & \textbf{40.3} & \textbf{9.1} & \textbf{49.9} & \textbf{55.0} & 51.8          & 85.9          & 88.4          & 73.7          & \textbf{47.3} & 87.1          & 58.1          & 53.0          & 61.0          & 60.4          & 66.7          \\
Ours (Cosine) & 86.3          & 44.2          & \textbf{88.3} & 39.2          & 7.5          & 49.2          & 54.7          & \textbf{54.7} & \textbf{87.2} & \textbf{90.7} & \textbf{73.8} & \textbf{47.3} & \textbf{87.4} & 55.9          & 53.7          & 60.7          & 61.3          & 68.1          \\
Ours (KL)     & \textbf{89.0} & \textbf{49.6} & 88.1          & \textbf{40.3} & 7.3          & 49.2          & 53.5          & 52.1          & 87.0          & 88.0          & \textbf{73.8} & 46.4          & 87.1          & \textbf{58.7} & \textbf{53.9} & \textbf{61.7} & \textbf{61.6} & \textbf{68.4}
\end{tabular}%
}
\label{tab:synthia_results}
\end{table}

\begin{figure}
\begin{center}
\begin{tabular}{cccc}
Image & DAFormer & Ours & Ground Truth \\
\includegraphics[width=2.8cm]{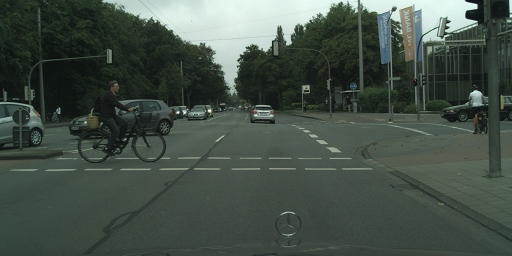} & \includegraphics[width=2.8cm]{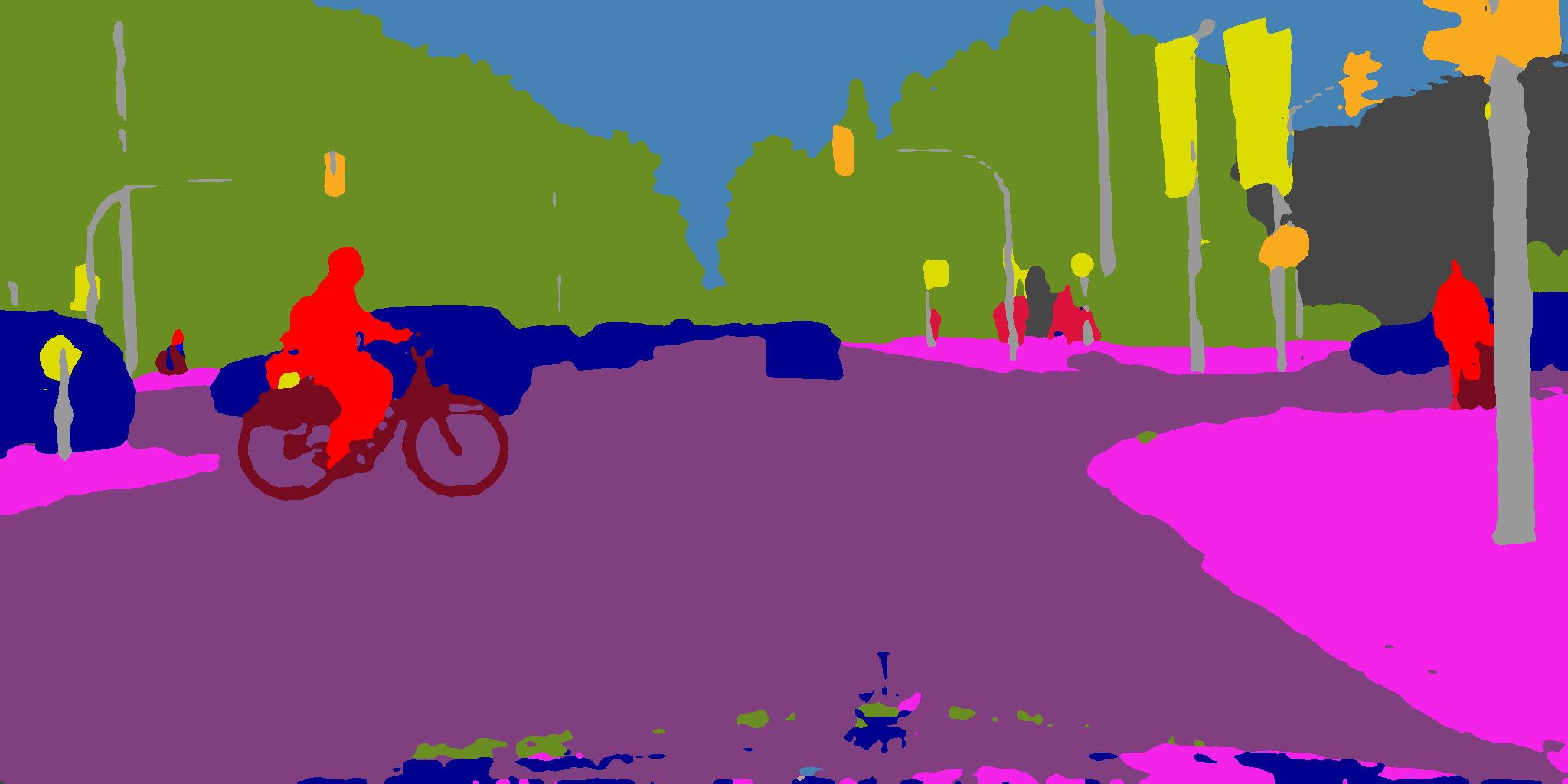} & \includegraphics[width=2.8cm]{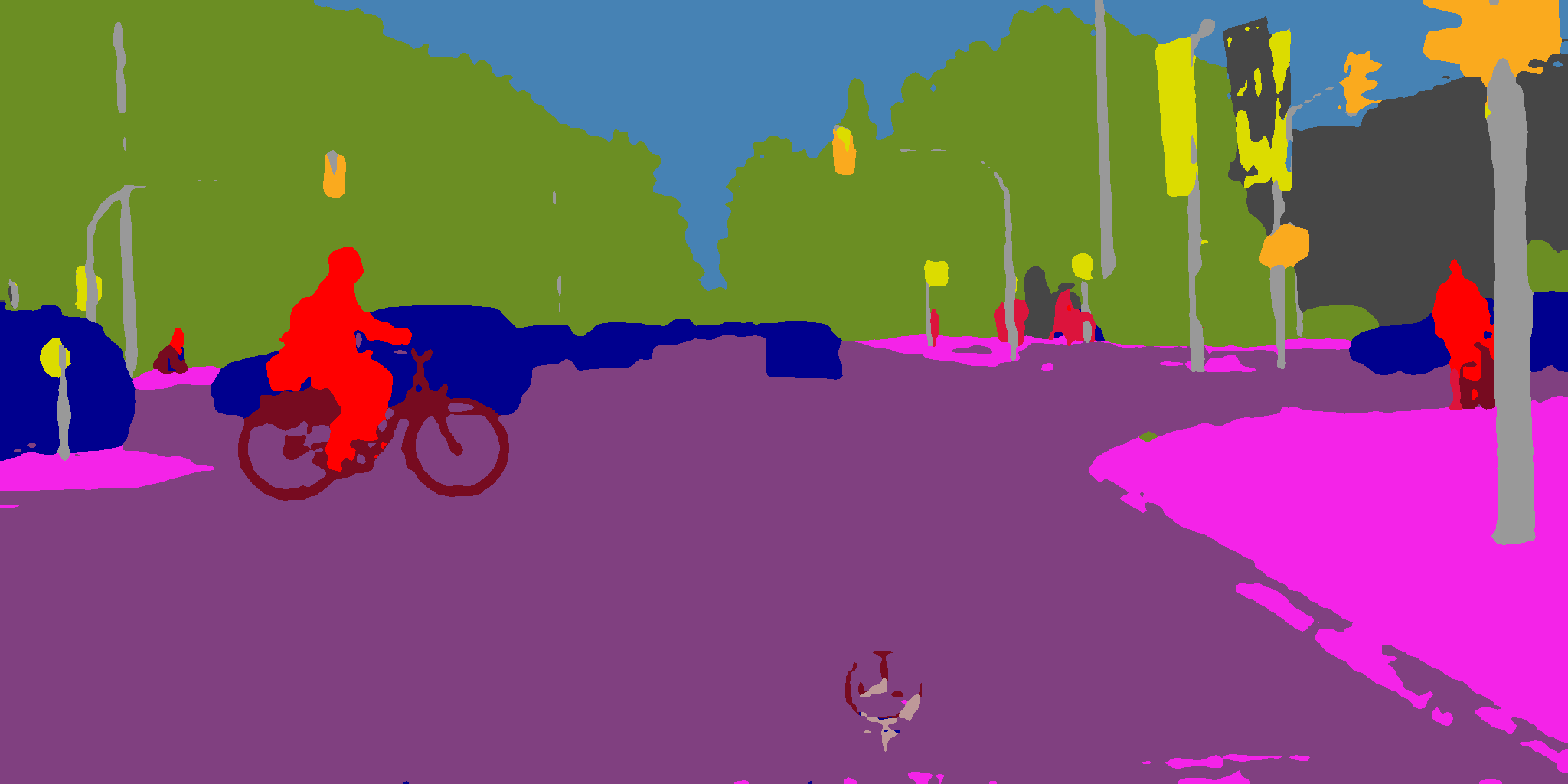} & \includegraphics[width=2.8cm]{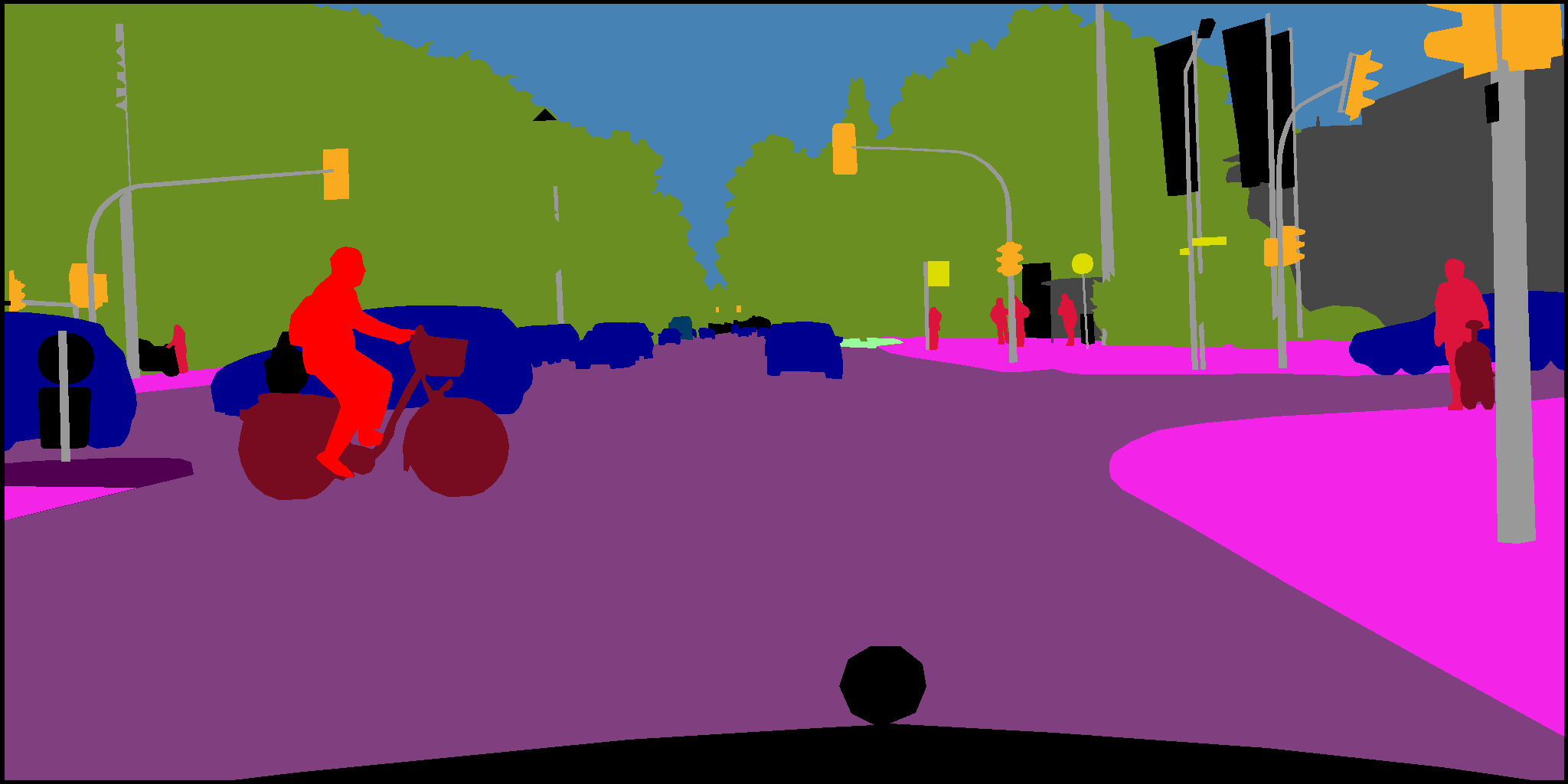} \\
\includegraphics[width=2.8cm]{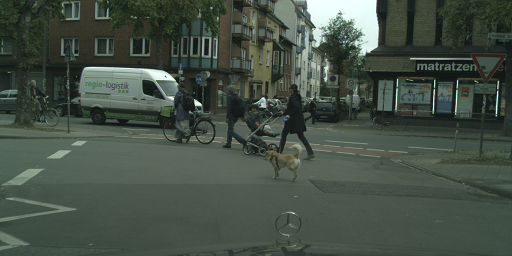} & \includegraphics[width=2.8cm]{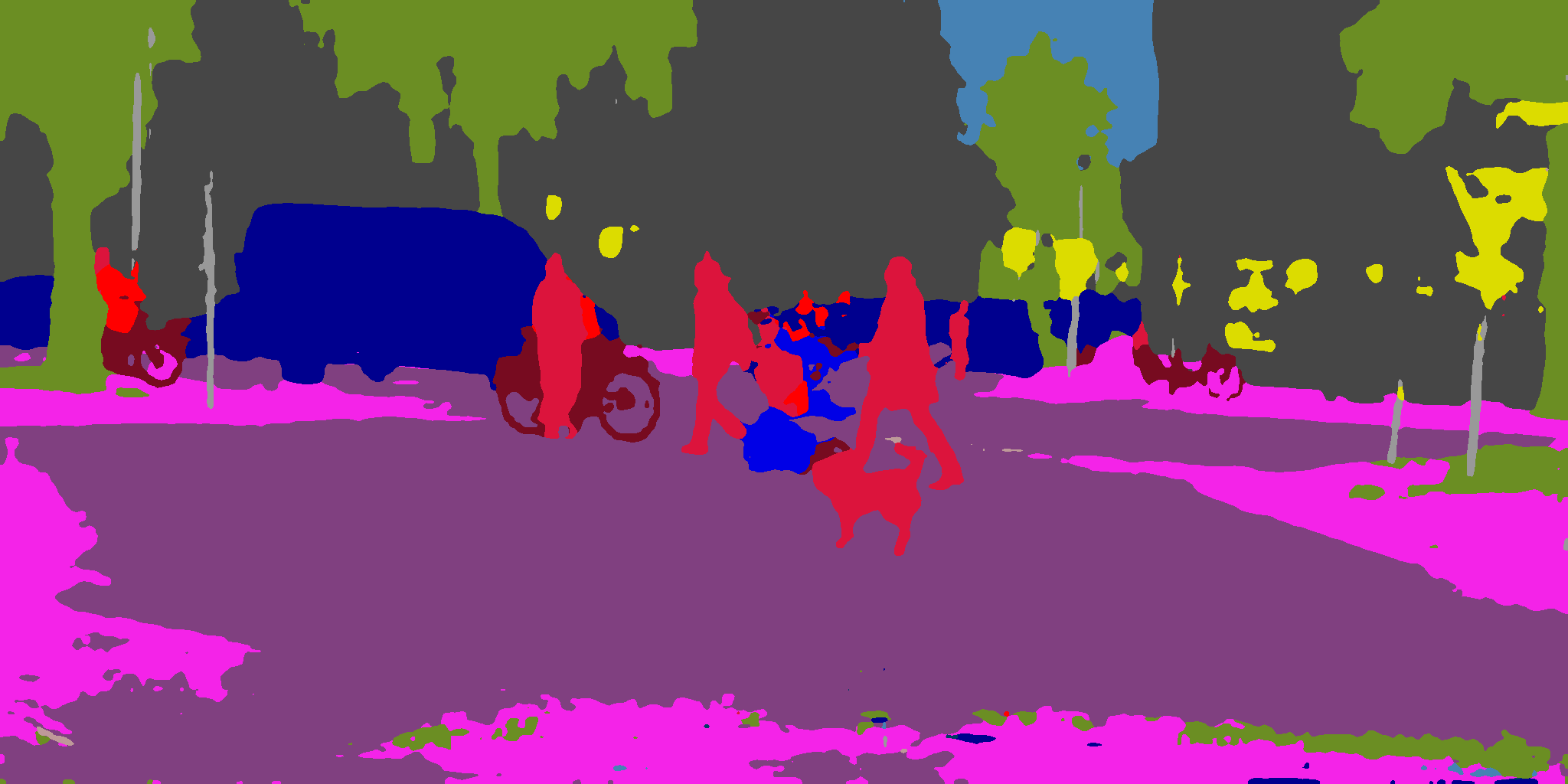} & \includegraphics[width=2.8cm]{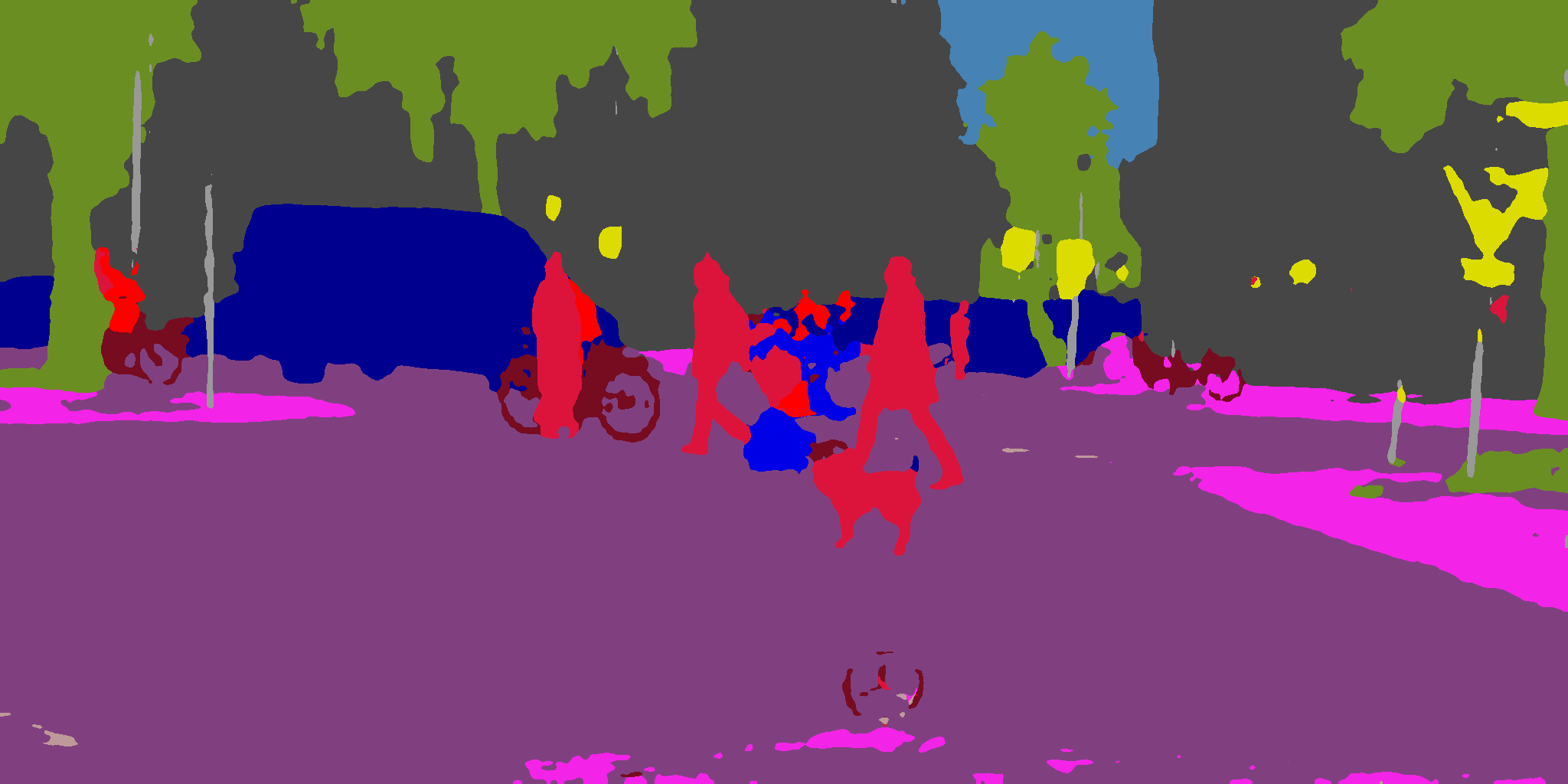} & \includegraphics[width=2.8cm]{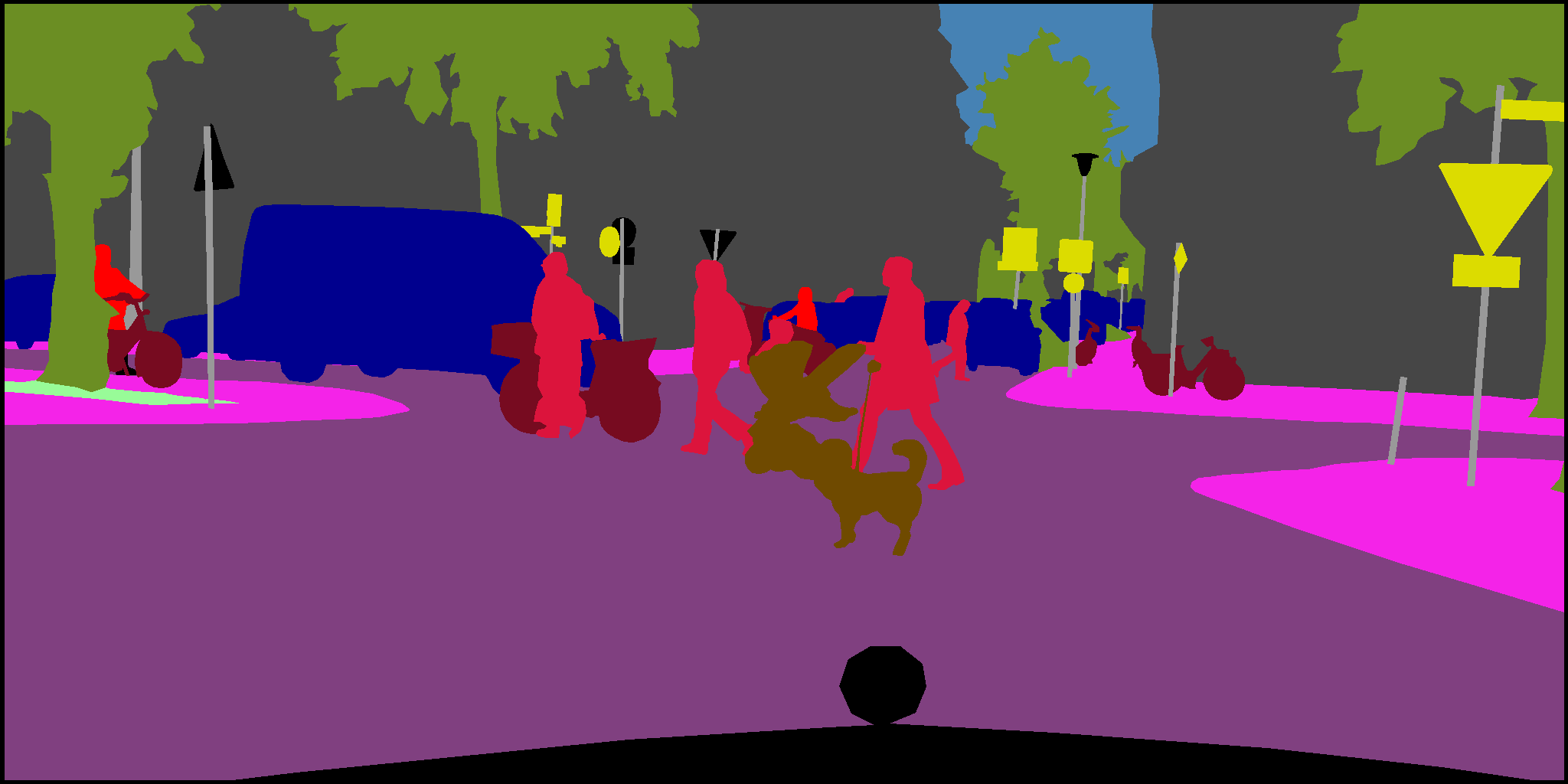} \\
\includegraphics[width=2.8cm]{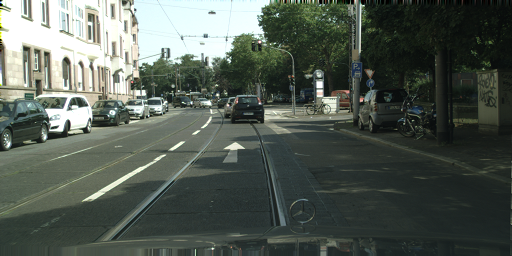} & \includegraphics[width=2.8cm]{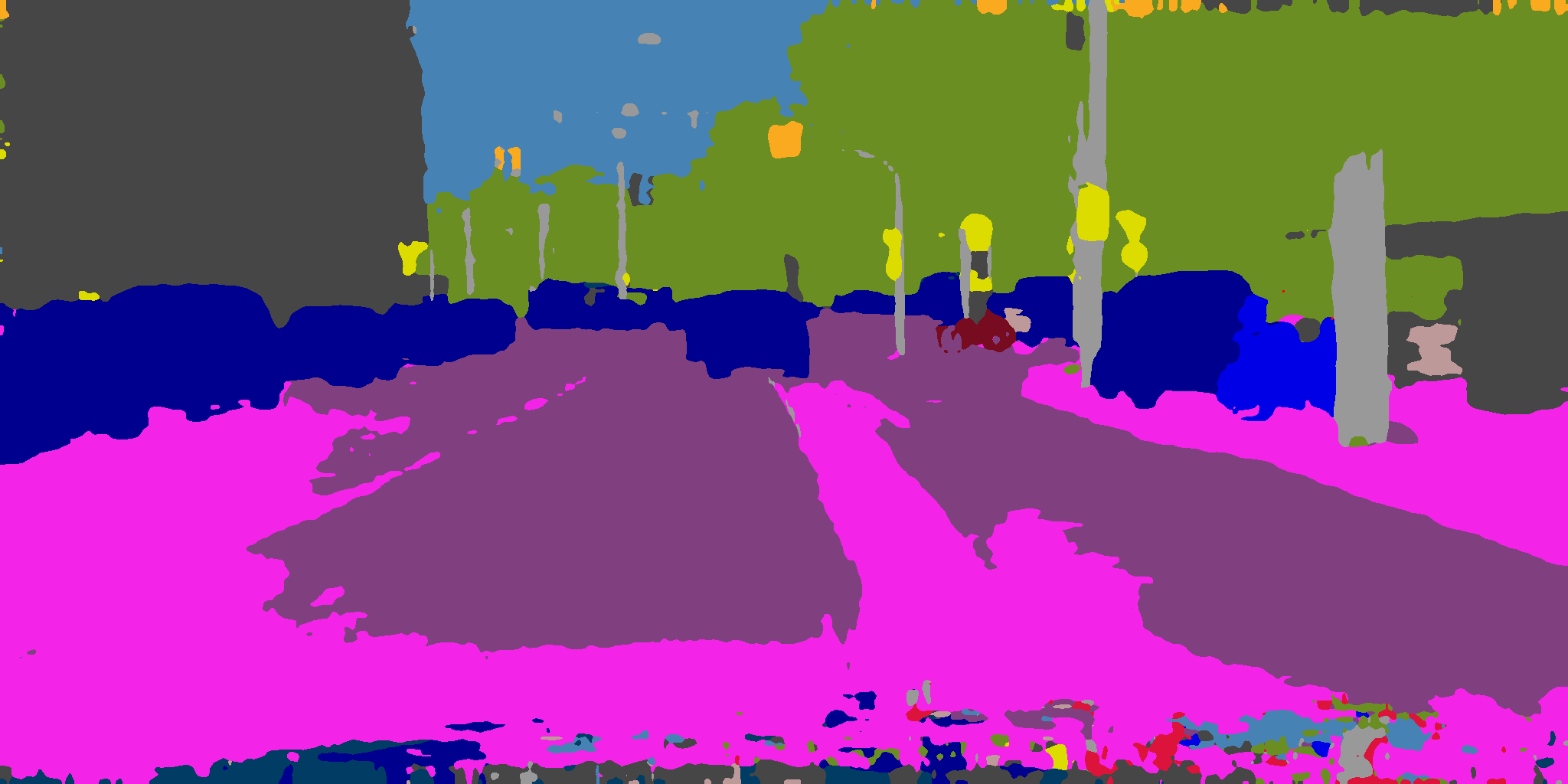} & \includegraphics[width=2.8cm]{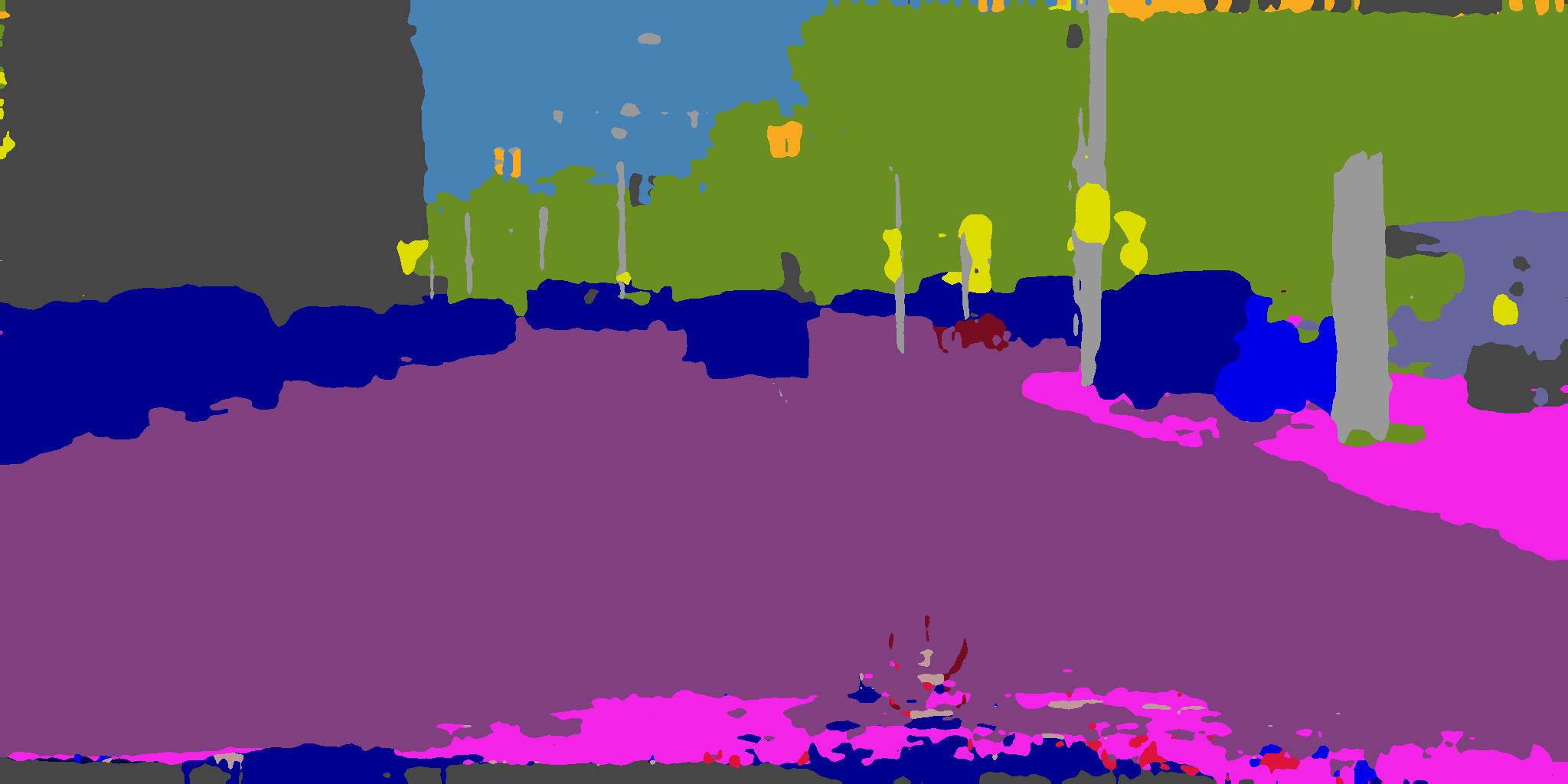} & \includegraphics[width=2.8cm]{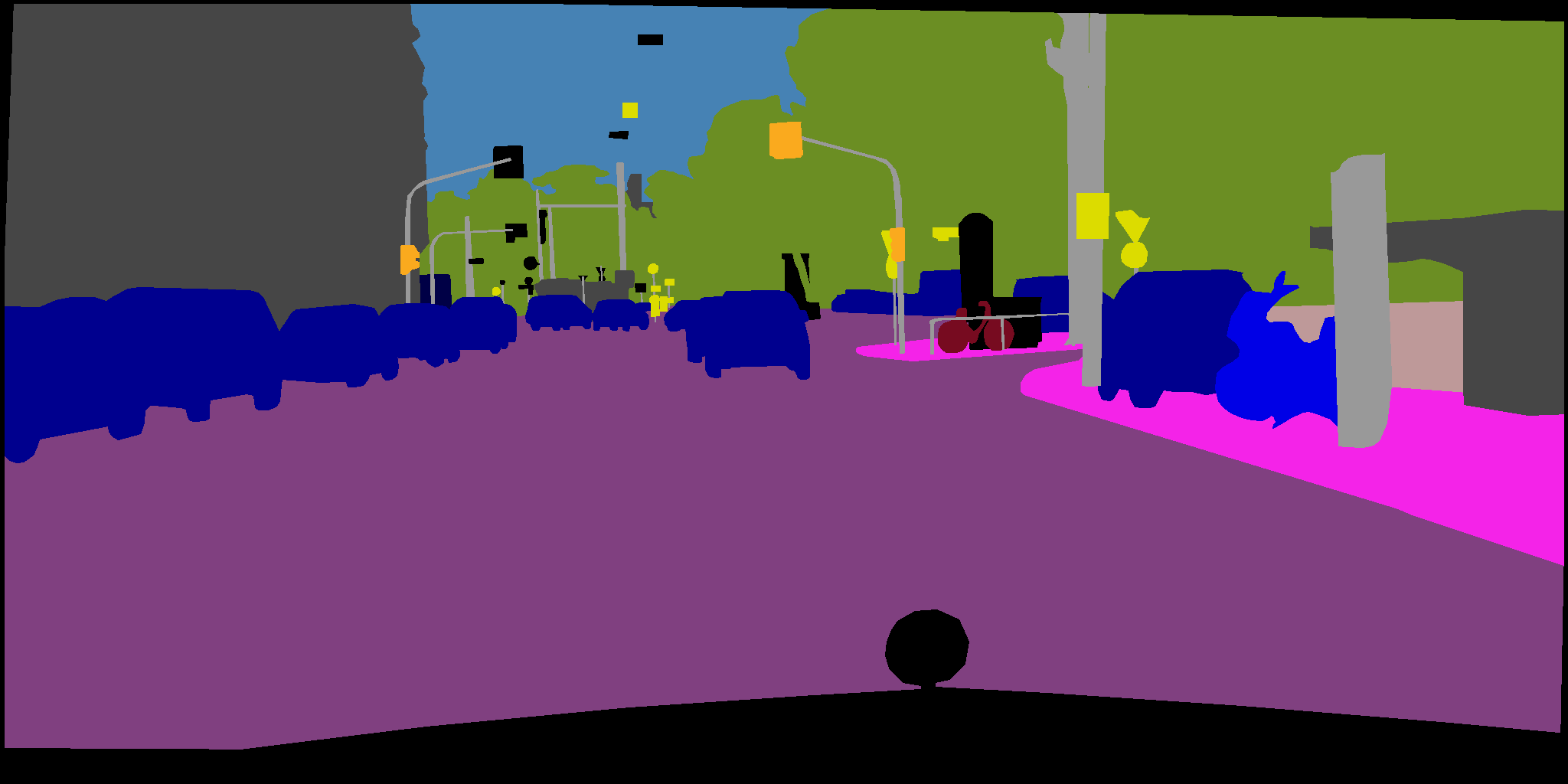} \\
\includegraphics[width=2.8cm]{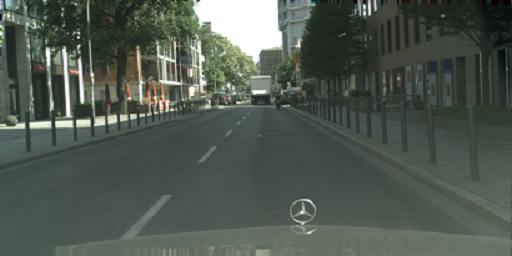} & \includegraphics[width=2.8cm]{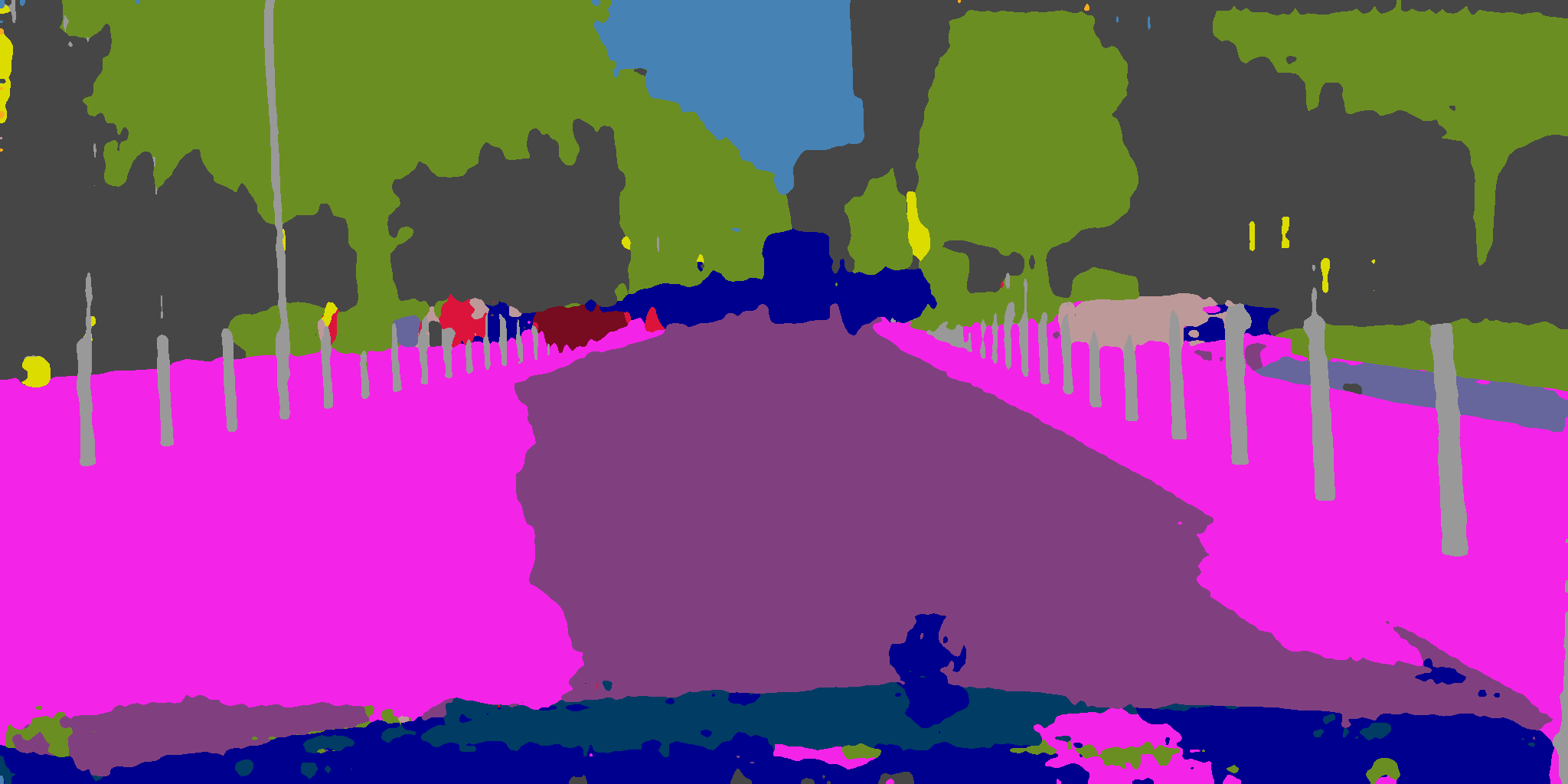} & \includegraphics[width=2.8cm]{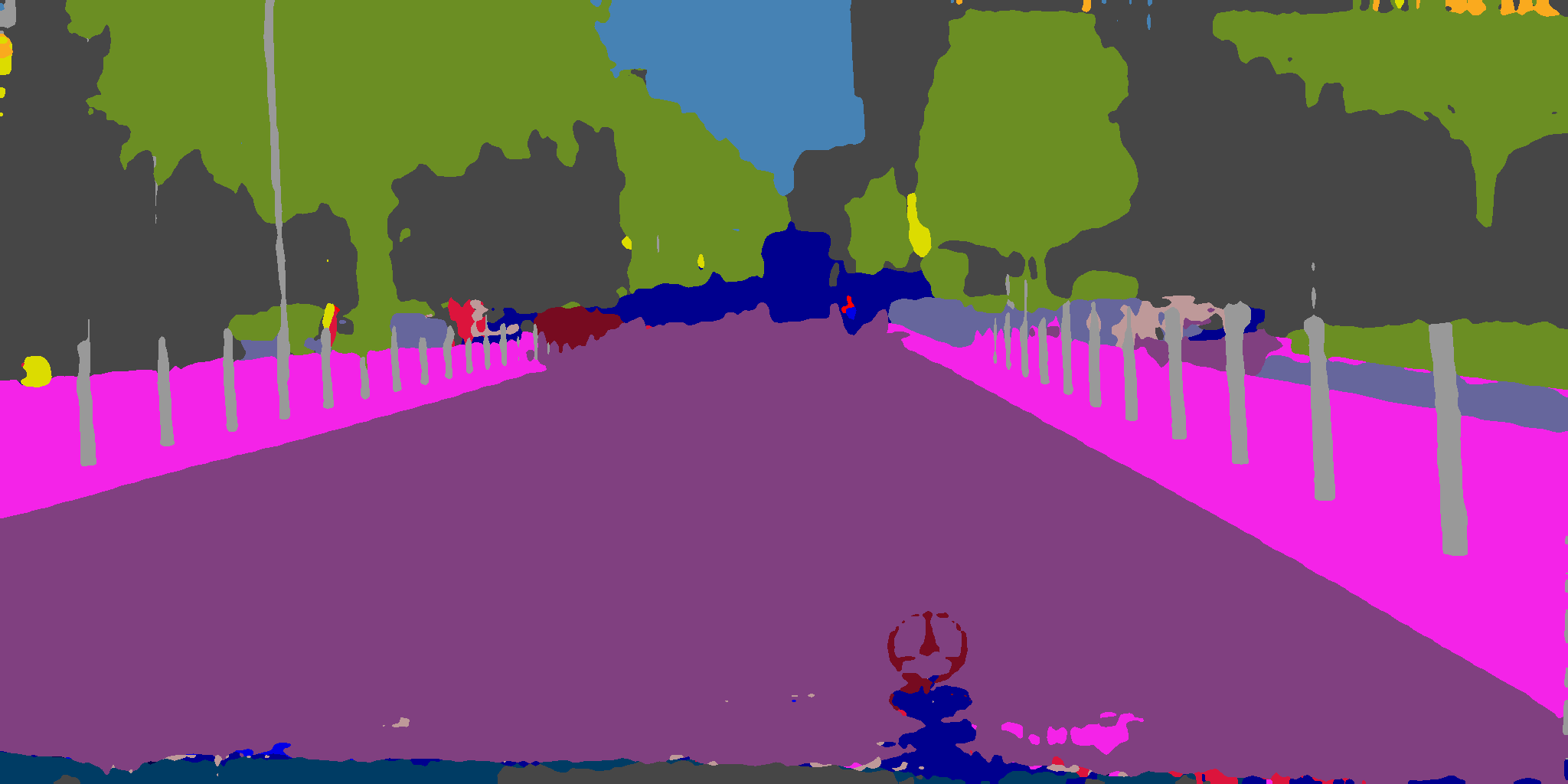} & \includegraphics[width=2.8cm]{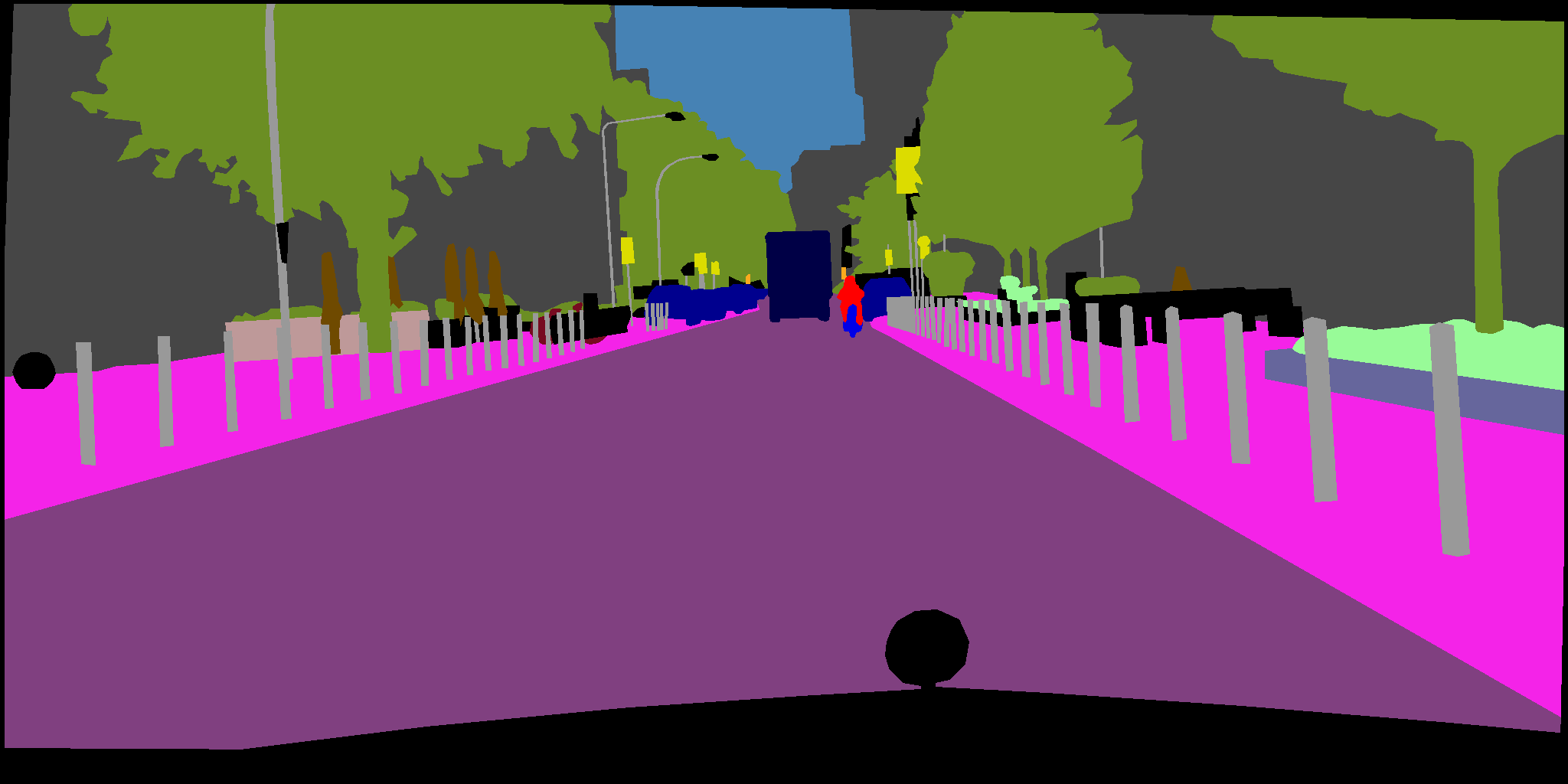} \\
\includegraphics[width=2.8cm]{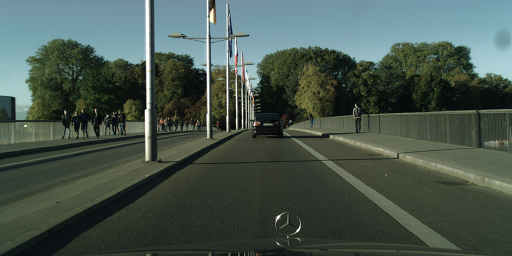} & \includegraphics[width=2.8cm]{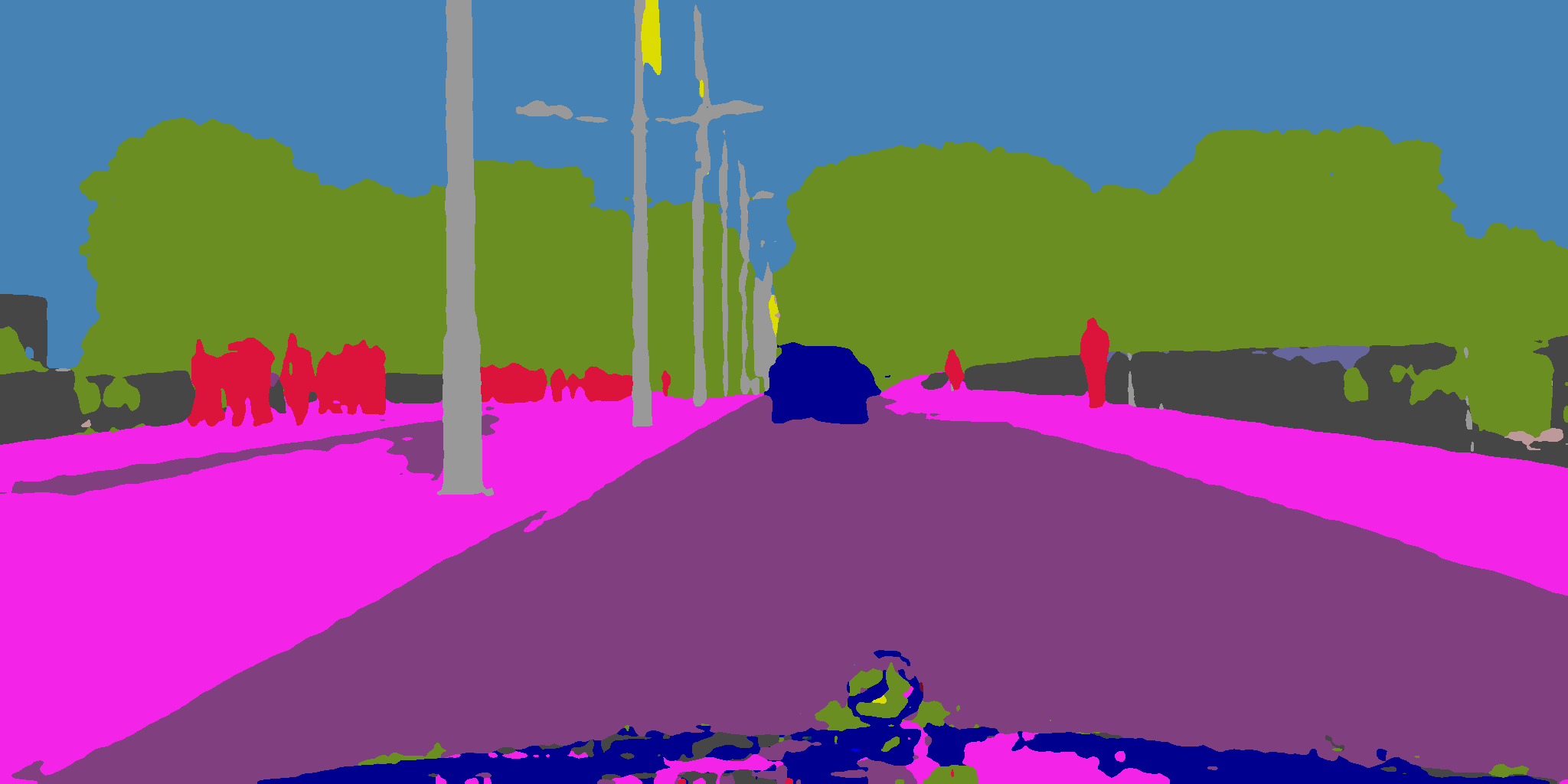} & \includegraphics[width=2.8cm]{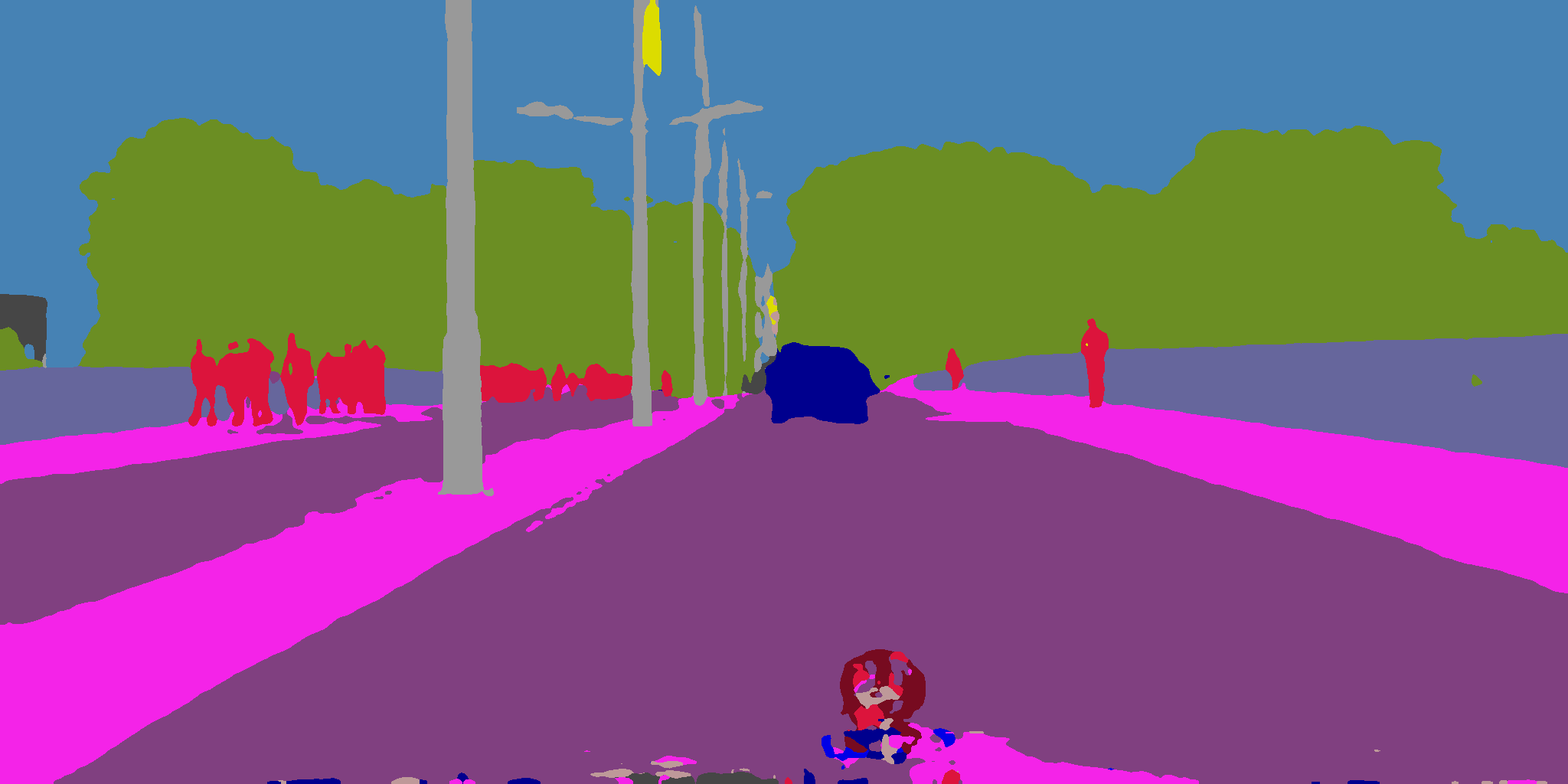} & \includegraphics[width=2.8cm]{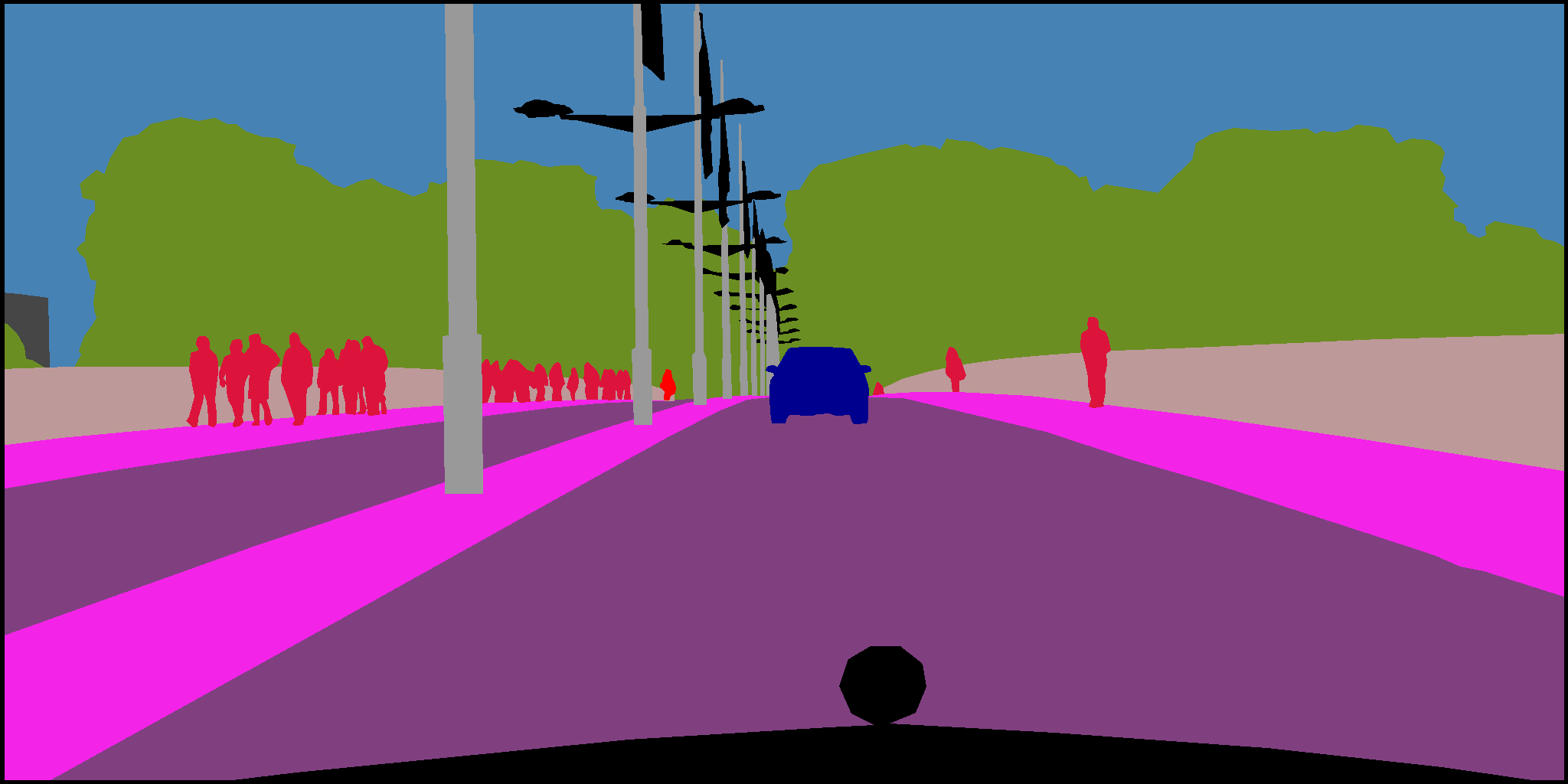} \\
\includegraphics[width=2.8cm]{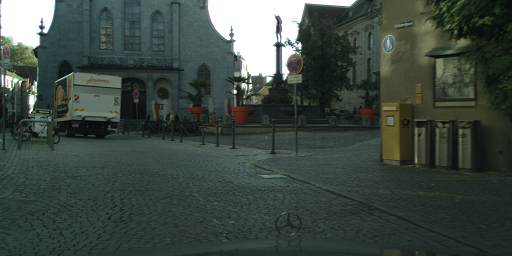} & \includegraphics[width=2.8cm]{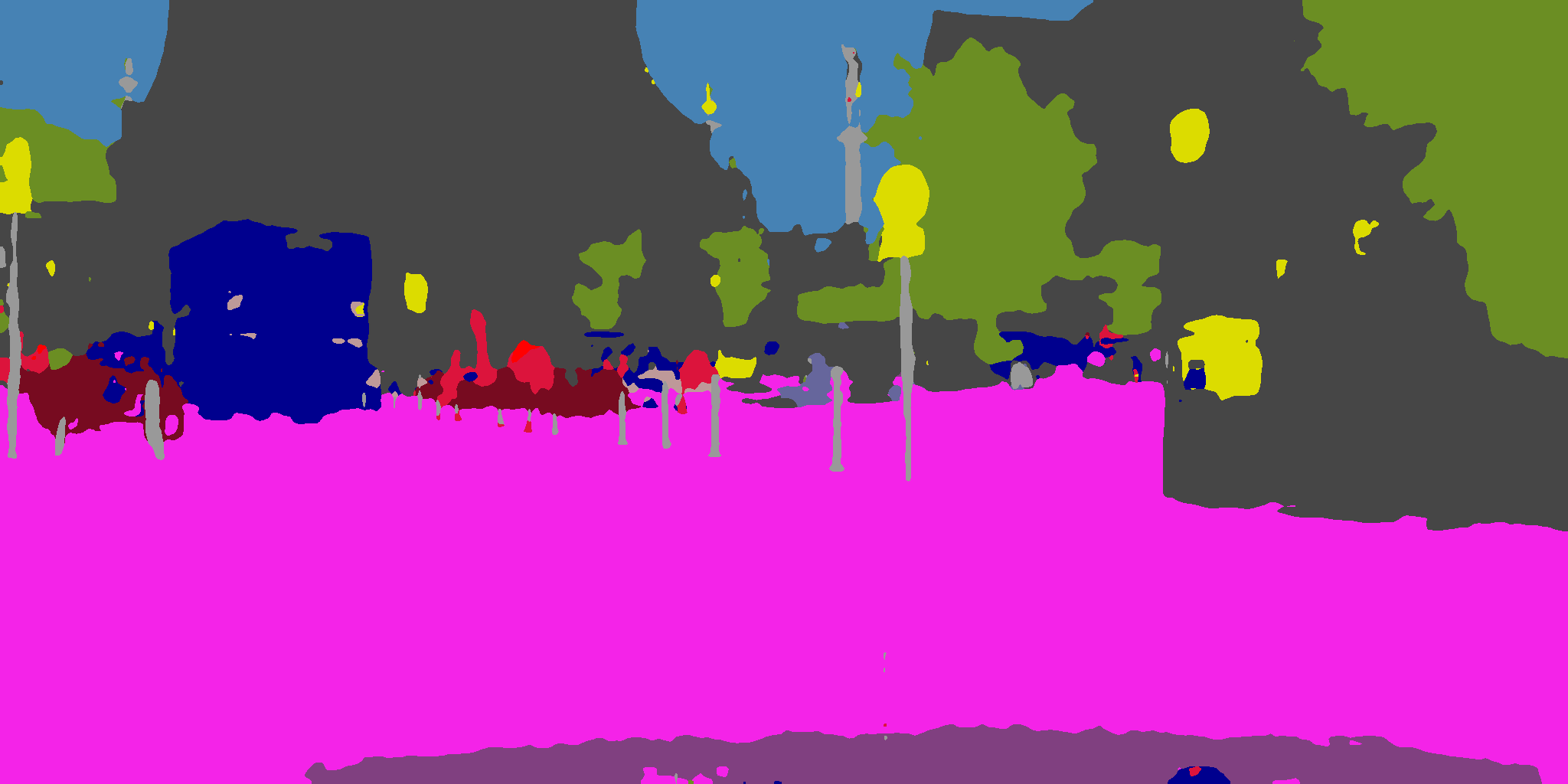} & \includegraphics[width=2.8cm]{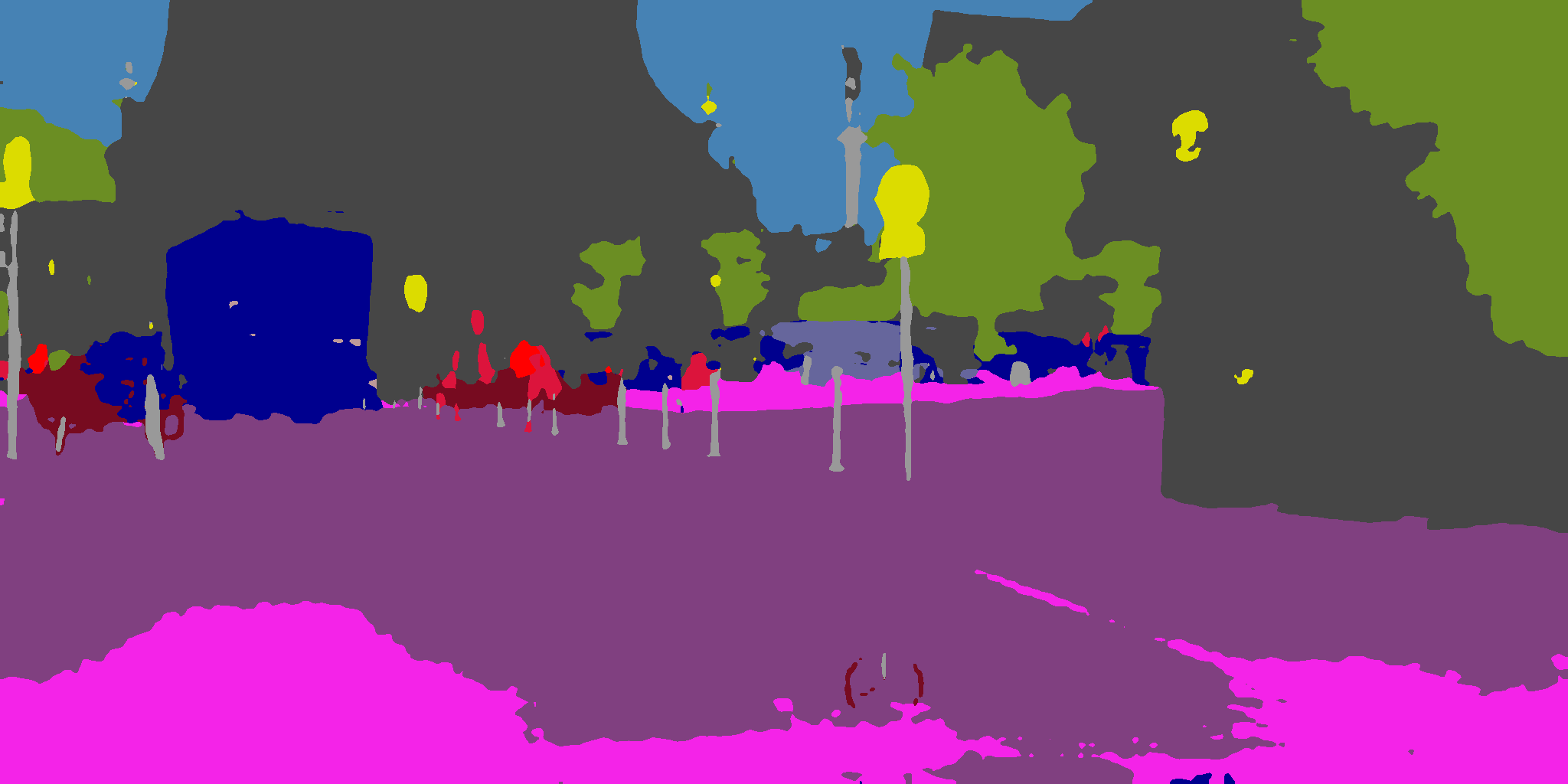} & \includegraphics[width=2.8cm]{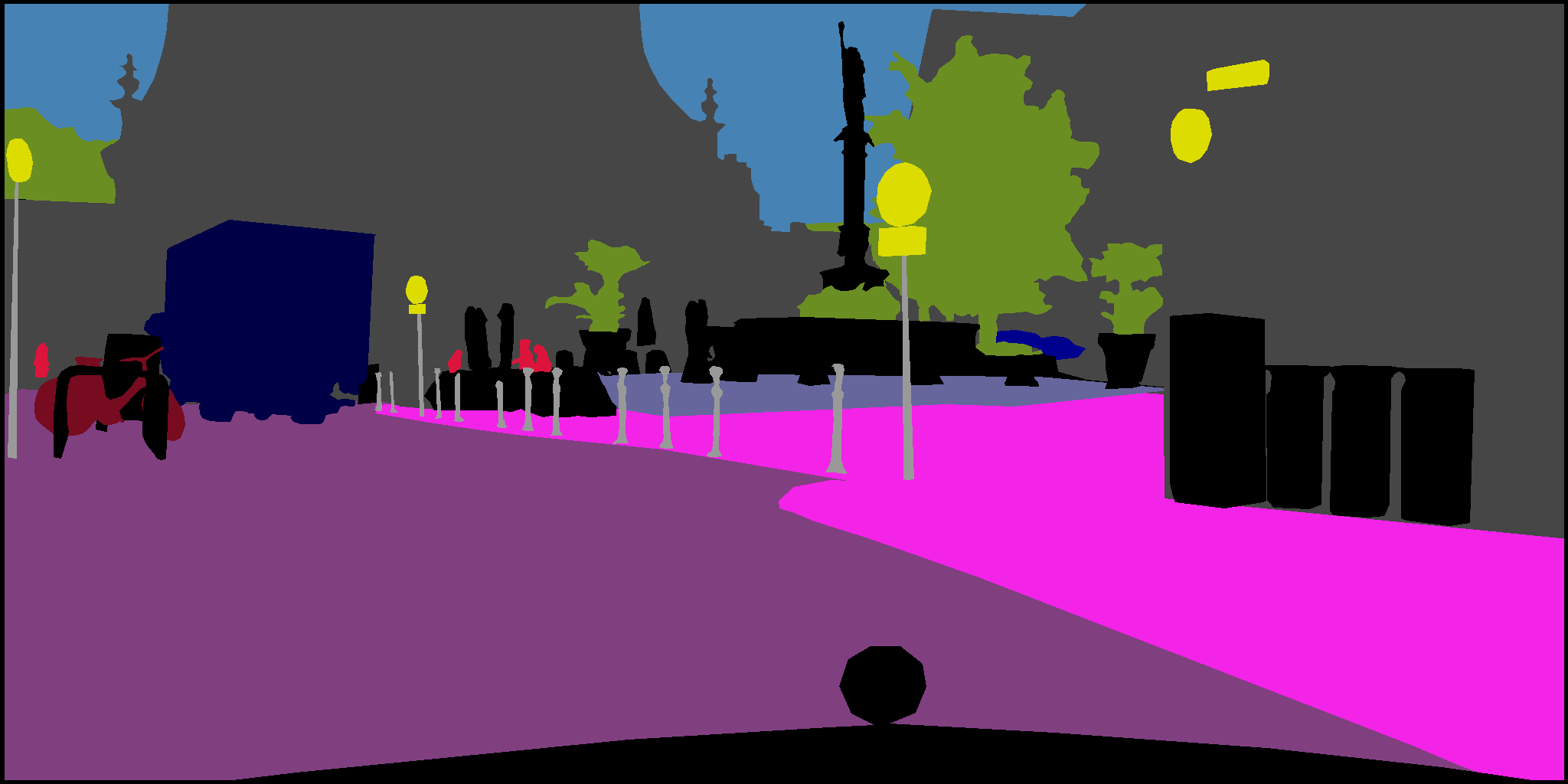} \\
\includegraphics[width=2.8cm]{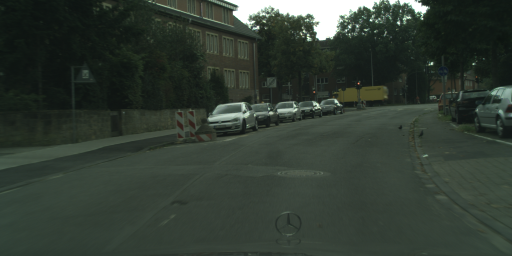} & \includegraphics[width=2.8cm]{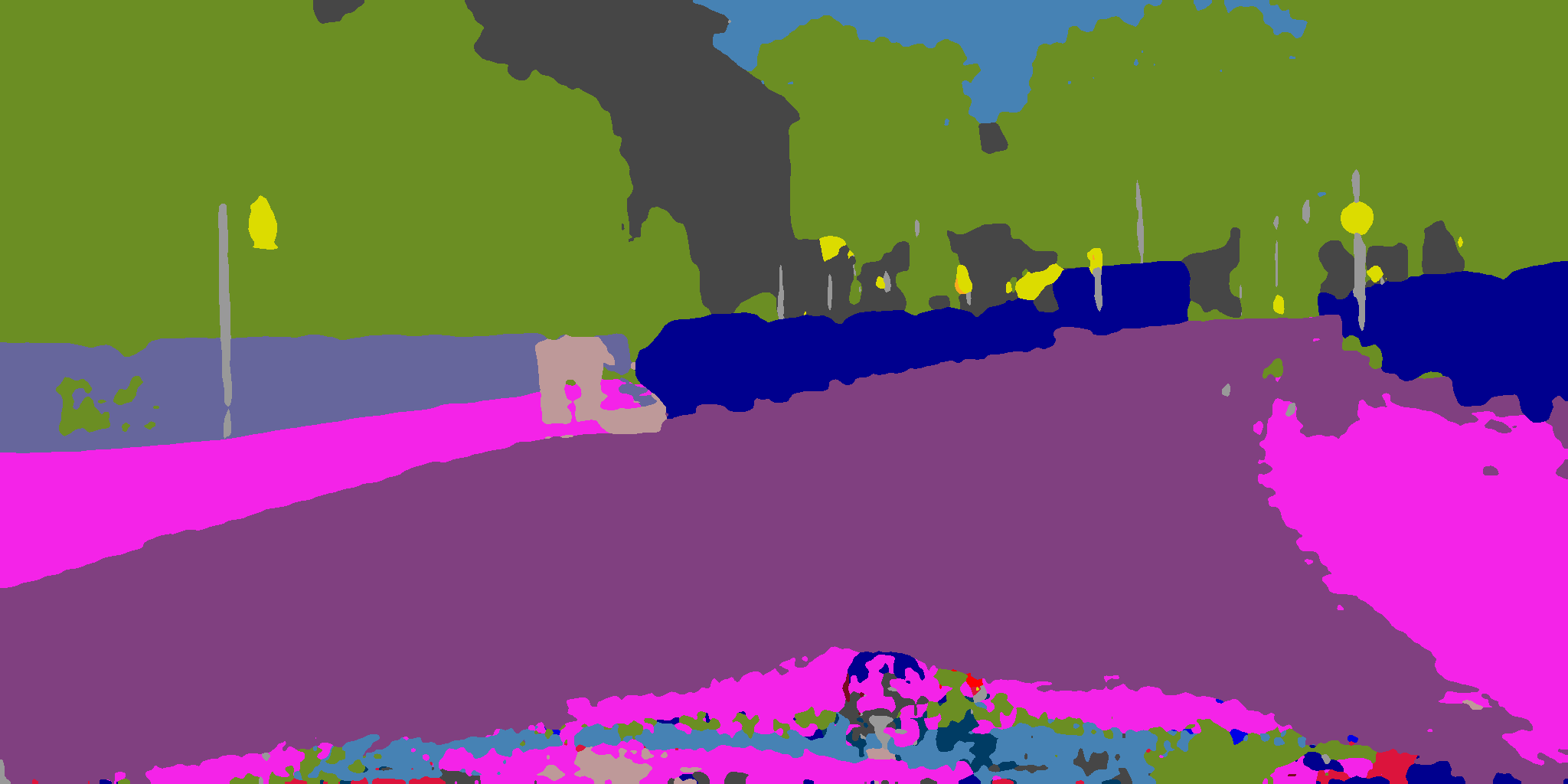} & \includegraphics[width=2.8cm]{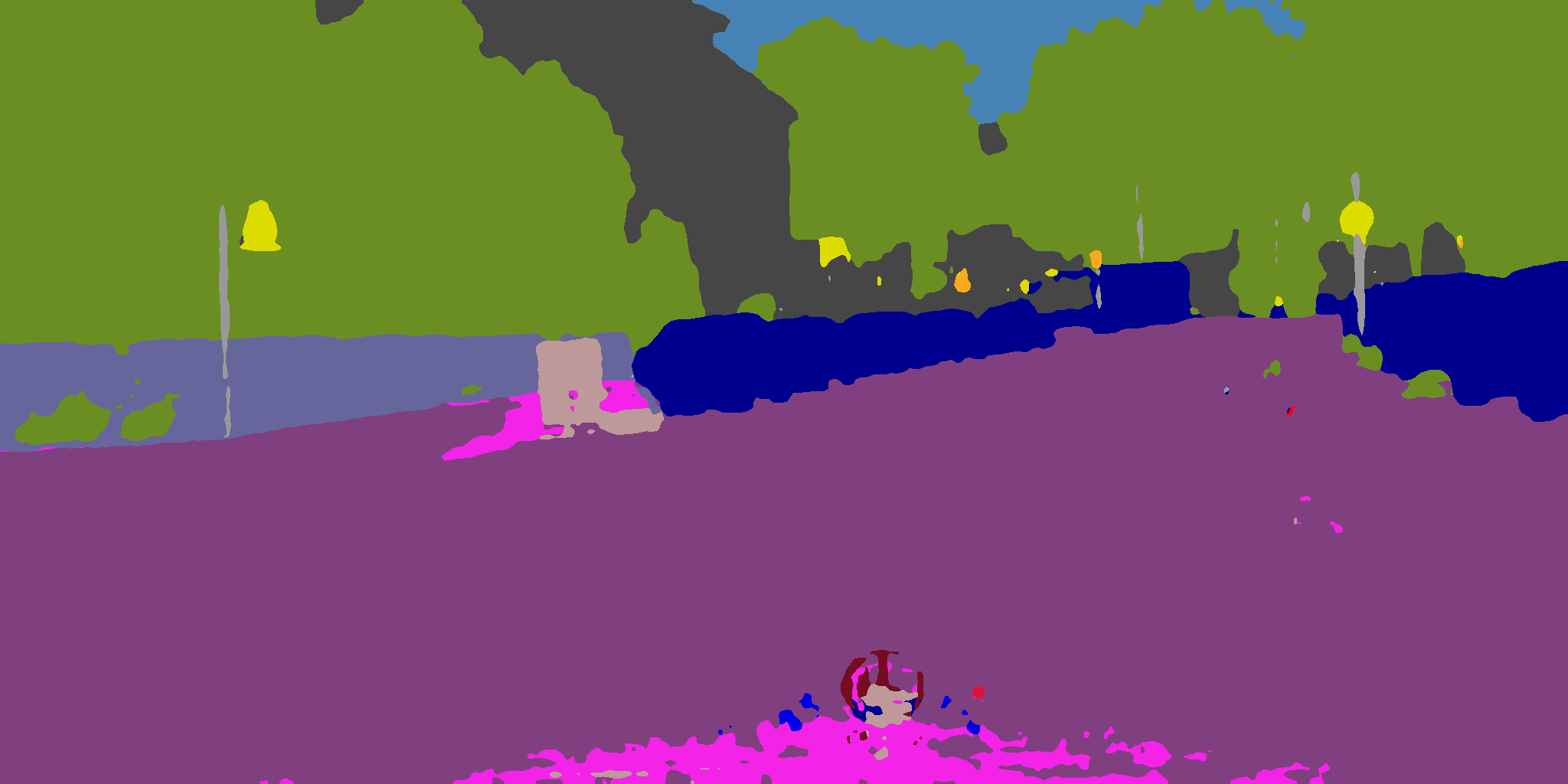} & \includegraphics[width=2.8cm]{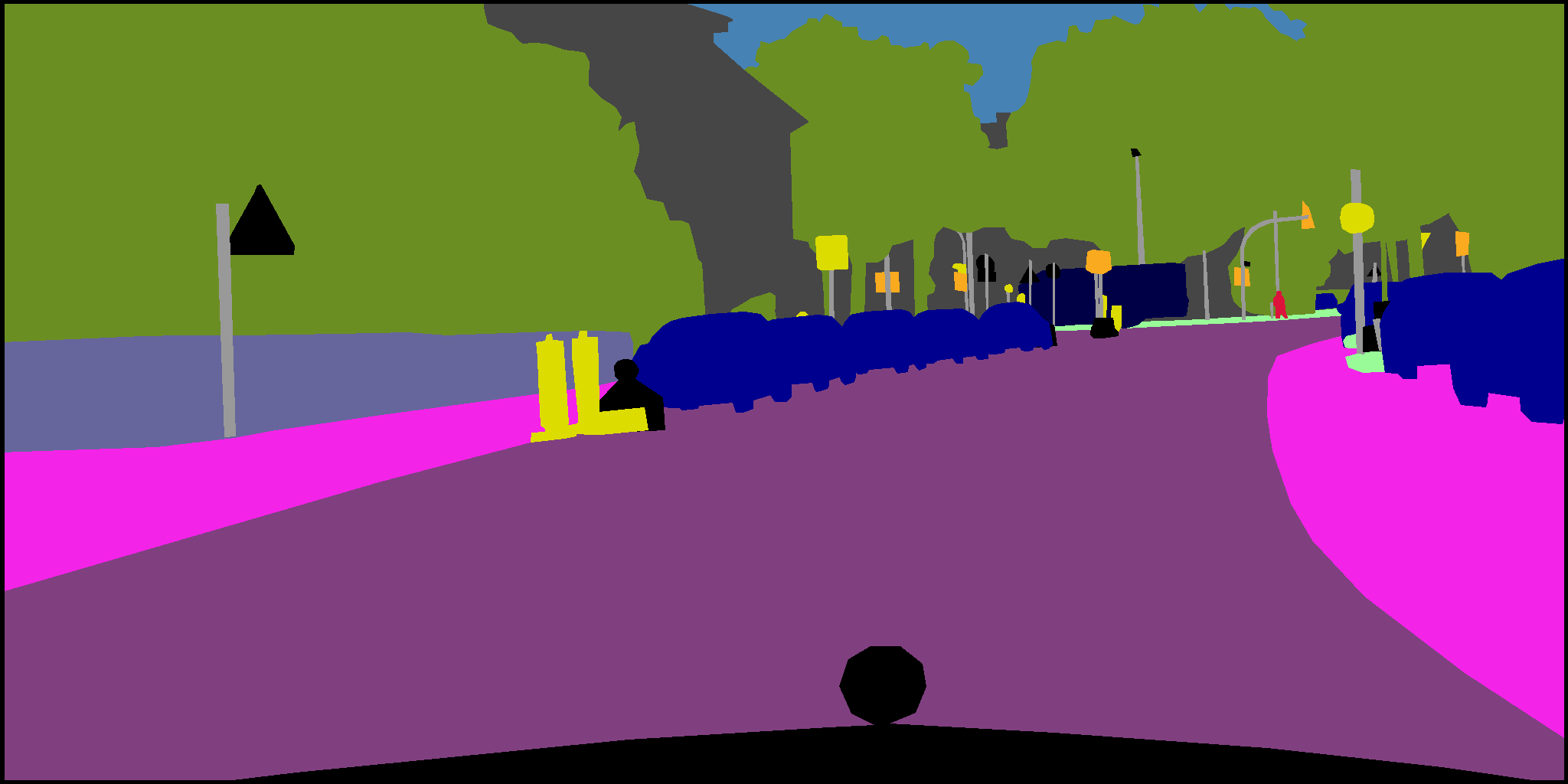}
\end{tabular}
\end{center}
\caption{Qualitative results comparing predictions on validation data of Cityscapes. From left: input image, predictions by DAFormer, predictions by our method, and the ground truth. The last row provides an example where DAFormer performed better compared to our method as it was able to correctly predict the sidewalks}
\label{fig:predictions}
\end{figure}

We compare the results of our method against other state-of-the-art UDA segmentation methods such as BDL~\cite{8954260}, ProDA~\cite{zhang2021prototypical} and DAFormer~\cite{DBLP:journals/corr/abs-2111-14887}.   
In table \ref{tab:gta_results}, we present our experimental results on the GTA5 to Cityscapes problem. It can be observed that our method improved UDA performance from an mIoU19 of 67.4 to 68.0 when cosine similarity is used in Equation~\ref{eq:similarity} and 68.2 when KL divergence is used. 
Table \ref{tab:synthia_results} shows our experimental results on the SYNTHIA to Cityscapes problem. Similarly, our method improved performance from an mIoU16 of 60.4 to 61.3 with cosine similarity and 61.6 with KL divergence.

\begin{figure}
\begin{center}
\begin{tabular}{ccc}
Target Image & DAFormer & Ours \\
\includegraphics[width=3.8cm]{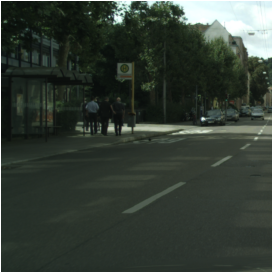} & \includegraphics[width=3.8cm]{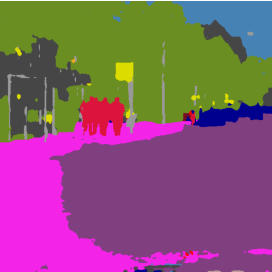} & \includegraphics[width=3.8cm]{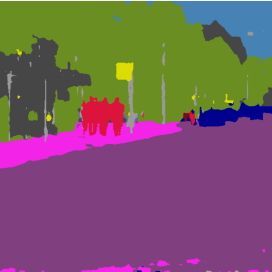} \\
\includegraphics[width=3.8cm]{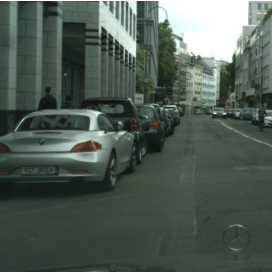} & \includegraphics[width=3.8cm]{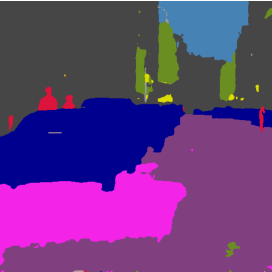} & \includegraphics[width=3.8cm]{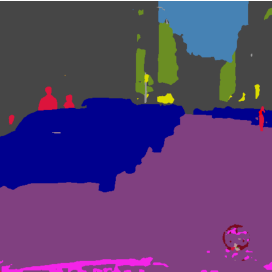} \\
\end{tabular}
\end{center}
\caption{Qualitative results on pseudo labels generated from the training data of Cityscapes. From left: target image, pseudo labels generated by DAFormer, and pseudo labels generated by our method using cosine similarity}
\label{fig:pseudo_labels}
\end{figure}

We also observed that our method was able to make notable improvements on the "Road" and "Sidewalk" categories. This is especially so on the SYNTHIA to Cityscapes problem, where we improved UDA performance on "Road" from 80.5 to 89.0 and "Sidewalk" from 37.6 to 49.6. We further verify this improvement in our qualitative analysis presented in Figure \ref{fig:predictions}, where we observed that our method had better recognition on the "Sidewalk" and "Road" categories. 
We attribute this improvement to our method's effectiveness in generating more accurate pseudo labels. We present pseudo labels generated during the training process in Figure \ref{fig:pseudo_labels}, where we observed more accurate pseudo labels for the "Road" and "Sidewalk" categories.

It should be noted that experimental results obtained using the DAFormer method in Tables \ref{tab:gta_results} and \ref{tab:synthia_results} were obtained by averaging 6 random runs using the official DAFormer implementation\footnote{\url{https://github.com/lhoyer/DAFormer}}. Even though we were unable to reproduce the exact numbers published in the DAFormer paper, we believe our experimental results for DAFormer are comparable. 

\subsection{Ablation Study}

\subsubsection{Number of Pixels Sampled}

\begin{table}
\caption{Influence of $N_{pair}$ on UDA performance. Results for all experiments were averaged over 3 random runs except for $N_{pair} = 512$, which was an average over 6 runs}
\label{tab:n_pair_results}
\begin{center}
\begin{tabular}{|c|c|c|c|c|c|c|} \hline
$N_{pair}$ & 4 & 16 & 64 & 256 & 512 & 1024 \\ \hline
mIoU16 & \textbf{61.4} & 60.8  & \textbf{61.4} & 61.1 & 61.3 & 60.6 \\
\hline
\end{tabular}
\end{center}
\end{table}

We conducted additional experiments on SYNTHIA to Cityscapes to observe the effect of $N_{pair}$ (from Equation \ref{eq:consistency-regularization-with-n-pairs}) on model performance. Theoretically, sampling more pixels for similarity calculation (i.e. a larger $N_{pair}$) allows us to have a more complete model of the inter-pixel relationship between predictions. However, empirical results in table \ref{tab:n_pair_results} suggests that $N_{pair}$ does not have significant influence on UDA performance. We observe that very small samples, such as $N_{pair} = 4$, were able to obtain comparable results with larger sample sizes.  

\begin{figure}
\begin{center}
\begin{tabular}{cccc}
\includegraphics[width=2.8cm]{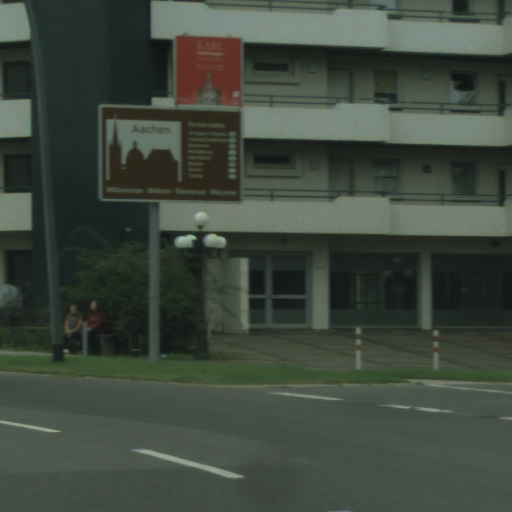} & \includegraphics[width=2.8cm]{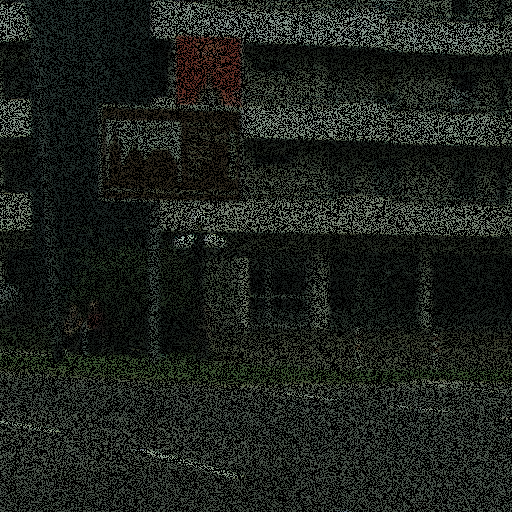} & \includegraphics[width=2.8cm]{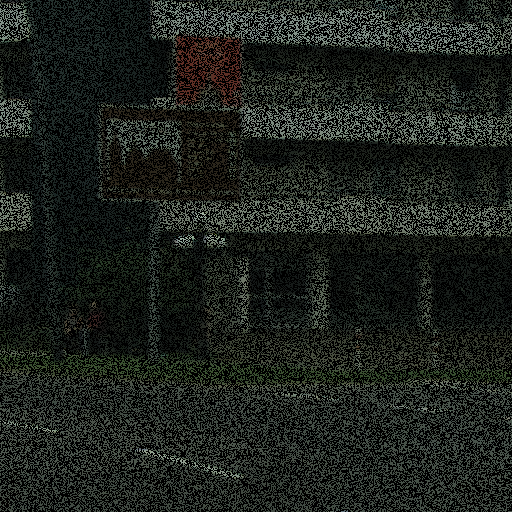} & \includegraphics[width=2.8cm]{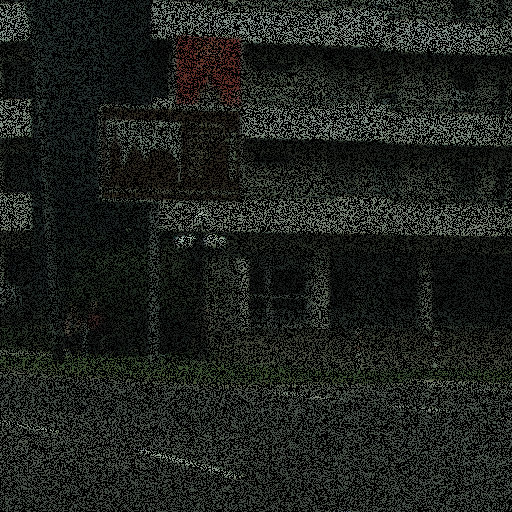} \\
(a) & (b) & (c) & (d)
\end{tabular}
\end{center}
\caption{Visualization of pixels sampled in experiments using $N_{pair} = 4$. (a) Target image cropped to $512 \times 512$; (b), (c) and (d) visualizes sampled pixels in three separate runs}
\label{fig:n_pairs_ablation}
\end{figure}

Additional experiments using $N_{pair} = 4$ were conducted to observe the locations of sampled pixels.
Visualization of our ablation study is presented in Figure \ref{fig:n_pairs_ablation}.
We found that after 40,000 training iterations, sampled pixels covered approximately 45.73\% of the 512$\times$512 images the network was trained on despite the small sampling size. This suggests that if a reasonable image coverage can be obtained during the training process, a small $N_{pair}$ is sufficient to model the inter-pixel relationship between predictions, allowing us to minimize computational cost of our consistency regularization method.
The influence of sampling coverage and sampling distribution on the effectiveness of consistency regularization is an interesting study that can be explored in the future. 

\subsubsection{Proximity of Sampled Pixels}

Kim et al. adopted cutmix augmentation \cite{Yun_2019_ICCV} in their consistency regularization method \cite{DBLP:journals/corr/abs-2001-04647} to limit sampled pixel pairs to within a local region. They theorized that pixel pairs that are in close proximity to each other have high correlation, and hence have more effect on UDA performance. We tested this theory on SYNTHIA to Cityscapes by performing $N_{box}$ crops and sampling $N_{pair}$ pixels from each crop. This localizes sampled pixels and restricts them to have closer proximity. Sampled pixels are then used to compute inter-pixel similarity to obtain a $N_{box} \times N_{pair} \times N_{pair}$ similarity matrix which is used for loss calculation in Equation \ref{eq:consistency-regularization}.
\begin{table}
\centering
\caption{Influence of $N_{box}$ and $N_{pair}$ on UDA performance. Total number of sampled pixels i.e. $N_{box} \times N_{pair}$ is kept at 512 for a fair comparison}
\label{tab:n_box}
\begin{tabular}{ccc|c}
Crop Size & $N_{box}$ & $N_{pair}$ & mIoU16 \\
\hline
256  & 32     & 16      & \textbf{61.2}   \\
128  & 32     & 16      & 61.1   \\
64   & 32     & 16      & 60.2  
\end{tabular}
\end{table}
We present the experimental results in Table \ref{tab:n_box} where three different crop sizes were varied to restrict the proximity of sampled pixels. We did not observe an improvement in UDA performance compared to results presented in Table \ref{tab:synthia_results}, suggesting that proximity of sampled pixels perhaps may not be that influential for consistency regularization.

\subsubsection{Measuring Inter-Pixel Similarity} \label{inter-pixel-similarity}

In Section \ref{consistency-regularization}, we adopted the method of Kim et al. to use cosine similarity in the measure of inter-pixel similarity \cite{DBLP:journals/corr/abs-2001-04647}. In this section, we conducted additional experiments on SYNTHIA to Cityscapes to observe the influence different methods of measuring inter-pixel similarity have on UDA performance. 

\begin{table}
\centering
\caption{Comparison of UDA performance using different methods to calculate inter-pixel similarity. We also provide the optimal $\lambda_c$ obtained using hyperparameter tuning}
\label{tab:similarity}
\begin{tabular}{lc|c}
Method            & $\lambda_c$   & mIoU16        \\
\hline
Cosine Similarity & 1.0         & 61.3          \\
Cross Entropy     & $1.0 \times 10^{-3}$ & 61.2          \\
KL Divergence     & $0.8 \times 10^{-3}$ & \textbf{61.6}
\end{tabular}
\end{table}

We tested the usage of cross entropy and KL divergence to measure inter-pixel similarity instead of cosine similarity in Equation \ref{eq:similarity}. Results from our empirical experiments are presented in Table \ref{tab:similarity}. We observed that all three methods provided comparable results with each other, with KL divergence providing slightly better results.

\section{Conclusion}
In this work we presented a new consistency regularization method for UDA based on relationships between pixel-wise class predictions from semantic segmentation models. Using this technique we were able to improve the performance of the state-of-the-art DAFormer method. We also observed that even with smaller number of sampled pixel pairs $N_{pair}$, this regularization method was still able to be effective. Therefore, with minimal computational cost, we are able to improve the results of self-training methods for unsupervised domain adaptation.

\noindent
\textbf{Acknowledgment}
This research is supported by the Centre for Frontier AI Research (CFAR) and Robotics-HTPO seed fund C211518008.

%
%
\bibliographystyle{splncs04}
\bibliography{domain_adaption}
\end{document}